\documentclass[sigconf]{acmart}

\usepackage{tikz}
\usepackage{amsthm}
\usepackage{amsmath}
\usepackage{bm}
\usepackage{thmtools,thm-restate}
\usepackage{enumitem}
\usepackage{pifont}
\usepackage{array}
\usepackage{float}
\usepackage{color}
\usepackage{caption}
\usepackage{multirow}
\usepackage{booktabs}
\usepackage{graphicx}
\usepackage{colortbl}
\usepackage{pythonhighlight}
\usepackage{hyperref}
\usepackage{comment}
\usepackage{lipsum}
\usepackage{bbm}
\usepackage{algorithm}
\usepackage[noend]{algpseudocode}
\usepackage{hyperref}
\usepackage{xcolor}
\usepackage{anyfontsize}
\usepackage{makecell}
\usepackage{mathtools}
\usepackage[T1]{fontenc}
\usepackage[font=small,labelfont=bf,tableposition=top]{caption}
\usepackage[normalem]{ulem}
\usepackage{subfig}
\usepackage{todonotes}

\hypersetup{
colorlinks=true,
linkcolor=cyan,
citecolor=red,
urlcolor=blue}
\usepackage{caption}

\DeclareMathOperator*{\argmin}{arg\,min}

\newtheorem{theorem}{Theorem}
\newtheorem{lemma}[theorem]{Lemma} 
 
\newtheorem{remark}[theorem]{Remark}

\newtheorem{definition}[theorem]{Definition}

\newcommand{\pluseq}{\mathrel{{+}{=}}}

\newcolumntype{P}[1]{>{\centering\arraybackslash}p{#1}}
\newcommand{\xingzhiswallow}[1]{{}}

\renewcommand{\paragraph}[1]{\textbf{\noindent{#1}}}

\newcommand*\colourcheck[1]{%
  \expandafter\newcommand\csname #1check\endcsname{\textcolor{#1}{\ding{52}}}%
}
\newcommand*\colourxmark[1]{%
  \expandafter\newcommand\csname #1xmark\endcsname{\textcolor{#1}{\ding{55}}}%
}
\colourcheck{blue}
\colourxmark{red}
\AtBeginDocument{%
  \providecommand\BibTeX{{%
    \normalfont B\kern-0.5em{\scshape i\kern-0.25em b}\kern-0.8em\TeX}}}

\newcommand{\supp}{supp}

\newcommand{\mc}{\mathcal{}}

\newcommand{\R}{\mathbb R}

\newcolumntype{L}[1]{>{\raggedright\let\newline\\\arraybackslash\hspace{0pt}}m{#1}}
\newcolumntype{C}[1]{>{\centering\let\newline\\\arraybackslash\hspace{0pt}}m{#1}}
\newcolumntype{R}[1]{>{\raggedleft\let\newline\\\arraybackslash\hspace{0pt}}m{#1}}

\setcopyright{acmcopyright}
\copyrightyear{2018}
\acmYear{2018}
\acmDOI{10.1145/1122445.1122456}

\begin{document}

\title[]{ Fast and Robust Contextual Node Representation Learning \\ over Dynamic Graphs }


\author{Xingzhi Guo}
\email{xingzguo@cs.stonybrook.edu}
\affiliation{%
  \institution{Stony Brook University}
  \city{Stony Brook}
  \country{USA}
}

\author{Silong Wang}
\email{23210980080@m.fudan.edu.cn}
\affiliation{%
  \institution{Fudan University}
  \city{Shanghai}
  \country{China}
}

\author{Baojian Zhou}
\email{bjzhou@fudan.edu.cn}
\affiliation{%
  \institution{Fudan University}
  \city{Shanghai}
  \country{China}
}

\author{Yanghua Xiao}
\email{shawyh@fudan.edu.cn}
\affiliation{%
  \institution{Fudan University}
  \city{Shanghai}
  \country{China}
}

\author{Steven Skiena}
\email{skiena@cs.stonybrook.edu}
\affiliation{%
  \institution{Stony Brook University}
  \city{Stony Brook}
  \country{USA}
}
\renewcommand{\shortauthors}{short authors}

\xingzhiswallow{

A framework for dynamic graph learning.
A Fast optimization formulated PPR Procedure.
A instantiate method, GoPPE.

There are generally two approaches in graph learning:  Network proximity-based methods (e.g., Deepwalk) and GNN-based methods (GCN/GAT), between them the most obvious difference is whether to use node's raw feature.

-- In the successful GNNs (GCN/Sage), the node $v_i$'s raw feature, $\bm x_i \in \mathbb{R}^{d}$, generally serves as message (spatial viewpoint) or graph signal (spectral viewpoint), propagating to its local neighbors $\operatorname{Nei{}}(v_i)$ defined by the explicit network structure $\mathcal{G}$. Meanwhile it locally \textit{convolutes} the incoming neighbors' messages as its hidden representation, denoted as $h_i$. 
Typically, a trainable convolutional kernel $\bm W^{(l)} \in \mathbb{R}^{h \times d}$ is realized as a permuatation-invariant function in a recursive manner, for example, $\bm h_i ^{(l+1)} = \sum_{j \in {Nei{}}(v_i)} \bm W^{(l)} \bm h^{(l)}_{i}$, $\bm h^{0}_i = \bm x_i$, where $l$ is the recursion level. The GNN models are usually trained in a supervised manner.

-- Meanwhile Network embedding algorithms (e.g., Deepwalk) purely depend on the network structure and capture the node proximity in low-dimensional vectors via different dimension reduction techniques (e.g, matrix factorization, random projection, hashing, ... ). 
For example, Single Source Personalized PageRank (\textit{SSPPR}) captures the node-wise proximity. Denote SSPPR as $\bm \pi_i \in \mathbb{R}^{|V|}$ where $V$ is the vertex set of the graph, one can further employ dimension reduction function \footnote{ e.g., random projection or hash-based reduction} $f: \mathbb{R}^{|V|} \rightarrow \mathbb{R}^{d}$, such that $\|  f(x_i) - f(x_j) \|_2^2 \leq (1+\epsilon)\| x_i -x_j \|_2^2 $

-- Because of the extra information encoded in node features, more trainable and usually supervised training,


1. Message passing suffers from over smoothing issue
2. Network proximity model

1. faster ista-based ppr calculation

2. application: PPR-based as positional encoding for 

\textcolor{red}{new idea: exploit warm start as tune $\epsilon$ or $\ell_1$-reg strength, making $\epsilon^{(0)} > \epsilon^{(1)} > \ldots >  \epsilon^{(k)}$.}

}

\begin{abstract}

Real-world graphs grow rapidly with edge and vertex insertions over time, motivating the problem of efficiently maintaining robust node representation over evolving graphs. 
Recent efficient GNNs are designed to decouple recursive message passing from the learning process, and favor Personalized PageRank (PPR) as the underlying feature propagation mechanism.  
However, most PPR-based GNNs are designed for static graphs, and efficient PPR maintenance remains as an open problem. 
Further, there is surprisingly little theoretical justification for the choice of PPR, despite its impressive empirical performance. 

In this paper, we are inspired by the recent PPR formulation as an explicit $\ell_1$-regularized optimization problem and propose a unified dynamic graph learning framework based on sparse node-wise attention.  
We also present a set of desired properties to justify the choice of PPR in STOA GNNs, and serves as the guideline for future node attention designs.  
Meanwhile, we take advantage of the PPR-equivalent optimization formulation and employ the proximal gradient method (ISTA) to improve the efficiency of PPR-based GNNs upto 6 times.  
Finally, we instantiate a simple-yet-effective model (\textsc{GoPPE}) with robust positional encodings by maximizing PPR previously used as attention.
The model performs comparably to or better than the STOA baselines and greatly outperforms when the initial node attributes are noisy during graph evolution, demonstrating the effectiveness and robustness of \textsc{GoPPE}.

\end{abstract}




\maketitle

\section{Introduction}
\label{sec:intro}

Real-world is changing all the time \cite{guo2022verba}. Specifically, many real-world graphs evolve with structural changes\cite{wen2022disentangled, hajiramezanali2019variational} and node attributes refinement over time. 
Considering Wikipedia as an evolving graph where Wiki pages (nodes) are interconnected by hyperlinks, each node is associated with the article's descriptions as attributes. 
For example, the node \textit{"ChatGPT"} initially
\footnote{The initial ChatGPT Wikipedia page: \url{https://en.wikipedia.org/w/index.php?title=ChatGPT&oldid=1125621134}} 
contained only had a few links redirecting to the nodes \textit{OpenAI} and \textit{Chatbot}, before being gradually enriched \footnote{The refined ChatGPT Wikipedia page shortly after the creation: \url{https://en.wikipedia.org/w/index.php?title=ChatGPT&oldid=1125717452}
}
with more hyperlinks and detailed descriptions.  Such examples motivate the following problem:
\begin{quote}
    How can we design an \textbf{efficient} and \textbf{robust} node representation learning model that can be updated quickly to reflect graph changes, even when node attributes are limited or noisy? 
\end{quote}

The proposed problem consists of two parts-- \textbf{1):} how to design a high-quality and robust node representation algorithm and \textbf{2):}  how to efficiently update the node representation along with the graph changes.  
%
%
%
%

One possible approach is through Graph Neural Networks (GNNs). Pioneered by SGC\cite{wu2019simplifying}, APPNP \cite{gasteiger2018predict}, PPRGo\cite{bojchevski2020scaling}, and AGP \cite{wang2021approximate},
they replace recursive local message passing with an approximate feature propagation matrix. For example, SGC formulates the node classification problem into $\bar{\bm y} = \operatorname{Softmax}(\bm S^{k} \bm X \bm \Theta)$ where $\bm S^k$ is the $k^{th}$ power of the normalized adjacency matrix,  $\bm X$ is the node attribute matrix and $\bm \Theta$ is the learnable classifier parameter.  This formulation decouples propagation from the learning process because $\bm S^k$ can be explicitly calculated independent of the neural model.

More importantly, this formulation provides a novel interpretation of GNN as a feature propagation process, opening doors to new propagation mechanism designs. 
In particular, Personalized PageRank (PPR) is a popular choice, and PPR-based models significantly improve GNN's efficiency and scalability. 
However, our interested problem is in dynamic settings. Although one can naively train multiple GNNs on every graph snapshot, the efficiency is compromised because they are designed for static graphs and have to re-calculate the propagation from scratch as the graph changes over time.

Recently, several works \cite{zheng2022instant,guo2021subset, guo2022subset, fu2021sdg} have adopted PPR in dynamic graph learning, and use incremental PPR maintenance algorithm \cite{zhang2016approximate} to update PPR as the graph evolves. 
Despite the fact that PPR is a very successful vertex proximity measure and provides great empirical performance, there is no clear formal framework to justify the choice of using PPR, and little theoretical guidance on the propagation mechanism design. 
In addition, the PPR-based GNN's efficiency largely depends on the PPR solver's speed. To our best knowledge,  all the previous works favored \textit{ForwardPush} as the underlying PPR method, which still has a large room  for improvement, especially under dynamic settings.

On the other hand, recent research \cite{fountoulakis2019variational,fountoulakis2022open} proposed a variational formulation of PPR as an explicit $\ell_1$-regularized quadratic optimization problem.  
Different from the traditional PPR interpretation as the random surfer model that can be solved algorithmically by power iteration\cite{page1999pagerank} and \textit{ForwardPush} \cite{andersen2006local}, the PPR-equivalent $\ell_1$-regularized optimization problem bridges two seemingly disjoint subjects (graph learning and optimization). 
One benefit is that this well-defined PPR-equivalent optimization problem can be solved efficiently by standard proximal gradient methods such ISTA and FISTA\cite{daubechies2004iterative, beck2009fast, chen2023accelerating, zhou2024iterative, bai2024faster}, which have great potential for more efficient GNNs.
Most importantly, the $\ell_1$-regularization reveals PPR's inherent sparse property, which is similar to the well-known \textit{LASSO} regression \cite{tibshirani1996regression} and online learning \cite{zhou2019dual, zhang2017fast, zhang2017fast2, zhang2016symmetrical, zhang2018dual} for robustness and interpretability.
While \textit{LASSO} keeps the most important subset of feature dimensions, the $\ell_1$-regularizer in PPR  enforces to only keep a subset of active or important nodes, especially in large graphs. Therefore, the resulting models are robust by only aggregating the most important \textit{local} features.

 \vspace{-3mm}
\begin{figure}[htbp]
    \centering
    \includegraphics[width=1.0\linewidth]{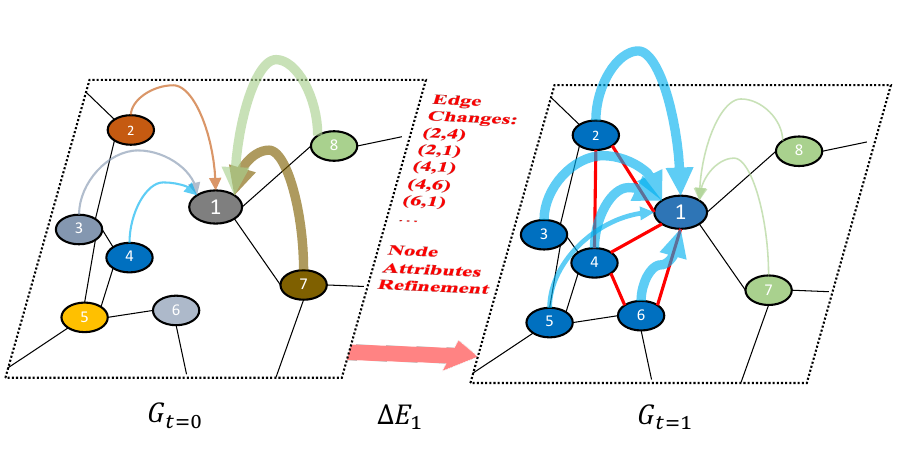}
    \vspace{-9mm}
    \caption{A cartoon illustrating contextualized node representation over a dynamic graph. 
    In the beginning, the graph has few edges  (black line) and relatively noisy node attributes (various colors). 
    As the graph evolves, the edges and node attributes become more complete, represented by red new edges and more consistent colored attributes.
    Accordingly, the node-wise attention (the arrows over nodes) for $u_1$ changes, favoring the nodes on its left side. 
    }
    \label{fig:goppe-overview}
\end{figure}

These findings inspire us to formulate PPR-based GNNs as a unified contextual node representation learning framework (shown in Figure \ref{fig:goppe-overview}) together with a new design of node positional encodings. 
Within this framework, we propose an elegant and fast solution to maintain PPR as the graph evolves and discuss several desired properties to justify the PPR choice with the theoretical ground. 
Finally, we instantiate a simple-yet-effective model (\textsc{GoPPE}) which has PPR-based node positional encodings, making the representation robust and distinguishable even when the node attributes are very noisy. 
The experiments demonstrate the potential of this framework for tackling the dynamic graph problem we are interested in.

%

We summarize our main contributions as follows: 
\begin{itemize}
    \item \textit{A novel contextual node representation framework for dynamic graphs} -- We propose a unified framework based on global node attention, which can explain the designs in the STOA propagation-based GNNs. 
    We identify the desired properties (e.g., attention sparsity, bounded $\ell_p$-norm, update efficiency) for the attention design, providing a guideline to justify the use of PPR and for designing the future attention-based GNN.  
    \item \textit{Improved efficiency of PPR-based dynamic GNNs through an optimization lens} -- Inspired by the new PPR optimization formulation, we improve the efficiency of PPR-based GNNs using the proximal gradient methods, specifically \textit{Iterative Shrinkage-Thresholding Algorithm} or \textit{ISTA}, demonstrating upto 6 times speedup in dynamic settings. 
    \item \textit{Instantiating \textsc{GoPPE}: a simple, effective and robust method} -- Within the proposed framework, we instantiate \textsc{GoPPE}, a simple-yet-effective method which is comparable in quality to and faster than the STOA baselines. Specifically, our novel PPR-based node positional encodings and the attention-based framework significantly outperform on noisy graphs, demonstrating its robustness against noise attributes during graph evolution.

\end{itemize}

The rest of this paper is organized as follows: Section \ref{sec:related-work}  clarifies different settings for dynamic graphs. 
Section \ref{sec:prelim}-\ref{sec:problem} discuss the preliminaries and problem formulation. 
We present our proposed node contextual learning framework in Section \ref{sec:method}, and discuss experimental results in Section \ref{sec:exp} .
Finally, we conclude and discuss future directions in Section \ref{sec:conclusion}.
\textit{We made our code accessible at \textcolor{blue}{\url{https://bit.ly/3ufpSQQ}} and will make it available upon publication.   }

\section{Related Work}
\label{sec:related-work}

Learning over dynamic graphs is an emerging topic, and has been studied from three perspectives with different problem settings: 

\paragraph{Co-evolution Models over Dynamic Graphs}: The problem of dynamic co-evolution focuses on the relations between node attributes and the growing network topology. 
For example, DynRep \cite{trivedi2019dyrep} is the pioneering model.
A typical task is to predict whether two vertices will be connected or interact in the future given the current graph topology and the node attributes.  
The key is to learn a node-wise relation model to represent the interaction probability among vertices, rather than updating the node embedding w.r.t. the graph changes. 
Due to the nature of $\mathcal{O}(|\mathcal{V}|^2)$ complexity, these models are not scalable when the node size is large.


\paragraph{Node Time Series Models over Dynamic Graphs}: 
Several temporal GNNs \cite{sankar2020dysat, rossi2020temporal,xu2020inductive} have been proposed for modeling the node embeddings w.r.t. the graph changes in \textit{discrete or continous time}. 
A typical example is \textsc{JODIE}\cite{kumar2019predicting}, which models the \textit{trajectory} of node embeddings across time and predicts its \textit{position} at a future timestamp. Similar to the co-evolution models, the essence is to obtain a compact embedding from the node history, and use it to extrapolate for future prediction. However, these models cannot update node embeddings quickly and usually involves data re-sampling and model fine-tuning while the graph changes.

\paragraph{Maintainable Node Embeddings Models over Dynamic Graphs}: Maintainable or incremental node embedding models focus on efficient embedding updates as the graph changes. 
Starting from the dynamic network embeddings which do not consider the node's raw attributes, previous works focus on online matrix factorization methods (e.g., incremental SVD \cite{brand2002incremental}), and incremental fine-tuning algorithms to adapt the static node embedding algorithms into dynamic (e.g., NetWalk\cite{yu2018netwalk}).  
However, they are not scalable because of the embedding dependency issue. For example, using the re-sample and fine-tune strategy,  one must update all node embeddings \textit{globally} to have one node embedding updated. 

Fortunately, \textit{localized} node embedding algorithms \cite{postuavaru2020instantembedding, guo2021subset, guo2022subset} adopt PPR which can be calculated for a subset of nodes individually with the complexity independent of node size. 
Meanwhile, one node's PPR is incrementally updatable and avoids calculating from scratch when the graph updates. 
Due to PPR's nature of local computation and maintainability, there have been follow-up works on the applications of efficient node anomaly tracking \cite{zhang2023subanom} and efficient feature propagation GNNs over dynamic graphs. 
Specifically, DynPPE \cite{guo2021subset} employs incremental \textsc{ForwardPush} algorithm to maintain PPR and apply local hashing method to obtain high-quality PPR-based embeddings, showing promising results in both node classification and anomaly tracking tasks.   
Another recent development is \textsc{InstantGNN} \cite{zheng2022instant}, which applies the same incremental PPR solver, but uses PPR as the  feature aggregation weights for GNNs design. 
Like its counterparts for static graphs (\textsc{PPRGo} \cite{bojchevski2020scaling} and \textsc{AGP} \cite{wang2021approximate}), the model obtains high-quality node representation while achieving high efficiency with the increment PPR algorithm. 

\paragraph{Our approach:}
\xingzhiswallow{Following this research line, we are curious about how to justify PPR as a good design choice for dynamic GNNs. Despite the fact that it can be interpreted as a specific GNN design with infinite layers, what are the desired properties (e.g., sparsity, locality) for one propagation mechanism to be suitable and efficient?  We seek to propose such properties within a contextual graph learning framework, justify PPR as a good choice and provide guidelines for future GNN designs.}
Following this research line, we investigate why PPR is a good design for dynamic GNNs, and what are the desired properties (e.g., sparsity, locality) for one propagation mechanism to be suitable and efficient.  
In addition, as PPR has been widely adopted with \textsc{ForwardPush} as its solver, we are curious about its efficiency under various dynamic settings (e.g., with large/small graph changes), and how to further improve GNN's efficiency. 
Finally, as the motivating scenario suggests, since the node attributes may be noisy during graph evolution, we aim to build a robust node representation against noisy attributes.  Specifically, we can re-use PPR (previously as the node attention) and convert it as node positional encodings with constant computation cost.

\begin{table}[ht]
\centering
\caption{Different STOA models can be realized within the proposed framework of various design choices. a : \textsc{PPRGo} only selects the top-$k$ PPR as attention weights.
b:  \textsc{InstantGNN} uses the generalized transitional matrix $\bm D^{\alpha} \bm A \bm D^{1-\alpha}$.
c: \textit{"S"} stands for Static PPR calculated from scratch in each snapshot. d: {\textit{"D"} means maintain PPR dynamically or incrementally.}
}
\vspace{-2mm}
\label{tab:design-compare}
\begin{tabular}{
p{0.10\textwidth} |p{0.05\textwidth}p{0.05\textwidth} p{0.06\textwidth}p{0.03\textwidth}p{0.08\textwidth}}
\toprule
 Method & Dynamic &  Attribute & Attention & PE & Solver \\
\midrule
 \textsc{InstantEmb.} & \redxmark & \redxmark & \redxmark & PPR & Push-S$^\text{c}$ \\
 \textsc{PPRGo} & \redxmark & \bluecheck & PPR$^\text{a}$ & \redxmark & Push-S \\
 \textsc{DynamicPPE} & \bluecheck & \redxmark & \redxmark & PPR & Push-D$^\text{d}$ \\
 \textsc{InstantGNN} & \bluecheck & \bluecheck & PPR$^{\text{b}}$ & \redxmark & Push-D  \\
\midrule
 \textbf{\textsc{GoPPE-\{S,D\}}} & \bluecheck & \bluecheck & PPR & PPR & ISTA-\{S,D\} \\
\bottomrule
\end{tabular}
\end{table}

\section{Notations and Preliminaries}
\label{sec:prelim}

\paragraph{Notations}: 
Given a undirected unweighted graph $\mathcal{G} = (\mathcal{V}, \mathcal{E}, \bm X)$ where $\mathcal{V}=\{1,2,\ldots,n\}$ is the node set,  $\mathcal{E} \subseteq \mathcal{V} \times \mathcal{V}$ denotes the edge set and $\bm X \in \mathbb{R} ^ { d \times |\mathcal{V}| }$ is the nodes' $d$-dim attributes.  
For each node $u$, we define $\operatorname{Nei}(u)$ to be the set of $v$'s neighbors; 
$\bm d$ denotes node degree vector where each element $d(i) = |\operatorname{Nei}(i)|$. 
The degree matrix is $\bm D := \operatorname{diag}(\bm d)$ and $\bm A$ is the adjacent matrix where $A_{i,j} \in \{0, 1\}$ and without self-loop. 
To simplify our notation, we use $d_i$ to denote the degree of node $i$. Let $\bm P = \bm D^{-1} \bm A$ be the random walk matrix and then $\bm P^\top = \bm A^\top \bm D^{-1}$ is the column stochastic matrix.


\subsection{Message Passing as Approx. Propagation}

In many successful and popular GNNs designs \cite{kipf2016semi, hamilton2017inductive}, the node attributes $\bm x_i $  serve as the message or graph signal, propagating to its local neighbors $\operatorname{Nei{}}(v_i)$ defined explicitly by the network structure $\mathcal{G}$. Meanwhile, $v_i$ locally \textit{convolutes} the incoming neighbors' messages as its hidden representation, denoted as $\bm h_i$. 
Typically, a set of trainable convolutional kernels $\bm W^{(l)}  \in \mathbb{R}^{h \times d} $ is used in a permutation-invariant function recursively ($l$ is the recursion level). For example, 
$\bm h_i ^{(l+1)} = \frac{1}{d_i} \sum_{j \in {Nei{}}(v_i)} \bm W^{(l)} \bm h^{(l)}_{j}$,
where $\bm h^{l=0}_i := \bm x_i$. The GNN models are usually trained in a supervised setting, learning local message aggregation and label prediction end-to-end. 

Interestingly, SGC \cite{wu2019simplifying} has simplified GNN design and decoupled the feature propagation from label prediction learning: 
\vspace{-1mm}
\begin{align}
    \bm H ^{(L)} &= \bm  W^{(L-1)} \bm H^{(L-1)} \bm P^{\top}, \bm P = \bm D^{-1} \bm A\nonumber\\
    \bm H ^{(L-1)} &= \bm  W^{(L-2)} \bm H^{(L-2)} \bm P^{\top} \nonumber\\
    \ldots \nonumber \\
    \bm H^{(1)} &= \bm W^{(0)} \bm H^{(0)} \bm P^{\top} \nonumber\\
    \bm H^{(0)} &:= \bm X \in \mathbb{R}^{d \times |\mathcal{V}|}, \text{ by definition.}\nonumber\\ 
    \bm H ^{(L)} &= \bm  W^{(L-1)} \bm  W^{(L-2)} \ldots \bm W^{(0)} \bm X \bm P^{\top} \ldots \bm P^{\top} \bm P^{\top} \nonumber\\
    & = \hat{\bm  W} \bm X \hat{\bm P}^{\top}, \text{ note that  } \hat{\bm  W} := \prod_{l=0}^{L-1} \bm W^{(l)},  \hat{\bm P} := \bm P^{L}, \label{eq:sgc}
\end{align}
where Equ.\eqref{eq:sgc} reveals strong connections between the GNN message-passing mechanism and well-studied vertex proximity measures in network science. 
\xingzhiswallow{
Specifically,  $\hat{\bm P} := (\bm D^{-1} \bm A)^{L}$ is the core component in power-method \cite{golub2013matrix} for problems like 
PageRank \cite{page1999pagerank, }. 
Let $\bm p_i$ is $i^{th}$ row of $\hat{\bm P}$, where $\| \bm p_i \|_{1} = 1$ and each element $\bm p_i[j]$ represents the  probability of a random surfer stops at $v_j$ from $v_i$ by taking $L$ times random walk, represnting a sense of proximity between $v_i$ and $v_j$.
As $L \rightarrow \infty$,  Equ.\ref{eq:sgc} essentially represents each node feature $\hat{\bm x}_i$ by \textit{Global Feature Aggregation}: $\hat{\bm x}_i = \bm X \hat{\bm P}^{\top}[:, i] = \sum_{j=0}^{|\mathcal{V}|} \bm p_i [j] \bm x_j$ where the probability vector $\bm p_i$ serves as aggregation weights.

}
\xingzhiswallow{
Alternatively, by taking into account 
$\hat{\bm  W} \bm X$ as a whole, 
Equ. \ref{eq:sgc} can be interpreted as \textit{global label smoothing} 
and each node is \textit{pre-classified} without accessing the graph structure: 
$\bm Y^{'} = \hat{\bm  W} \bm X$ where
$\bm Y^{'} \in \R^{|\mc{C}| \times |\mc{V}|}$, $\mc{C}$ is the label set, then smooth the labels according to the vertex proximity: 
$\bm Y[:,i] = \bm Y^{'} \hat{\bm P}^{\top}[:,i] = \sum_{j = 0} ^{|\mathbf{V}|} \bm p_i[j] \bm Y[:,j]$. 
Both interpretations give a \textit{contextualized} node representation for $v_i$ over $\mathcal{V}$. 
However, as $L \rightarrow \infty$, 
$(\bm D^{-1} \bm A)^{L}$ will converge to its first eigenvector (stationary property), making the output less distinctive and causing the over-smoothing issue \cite{Oono2020Graph}. 
}
Alternatively, by taking into account 
$\hat{\bm  W} \bm X$ as a whole, 
Equ. \ref{eq:sgc} can be interpreted as \textit{global label propagation}.
Both interpretations give a \textit{contextualized} node representation for $v_i$ over $\mathcal{V}$. 
However, as $L \rightarrow \infty$, 
$(\bm D^{-1} \bm A)^{L}$ will converge w.r.t. its first eigenvector (stationary property), making the output less distinctive and causing the over-smoothing issue \cite{Oono2020Graph}.

\paragraph{PPR as a node proximity design for GNNs}:  The celebrated algorithm Personalized Pagerank (PPR) provides a successful \textit{localized} design of node proximity.
One intriguing property of PPR is the \textit{locality} governed by the teleport component, making PPR quantities concentrated \textit{around} the source node. 
Starting from APPNP \cite{gasteiger2018predict} , PPR became a popular GNN design choice and empirically achieves outstanding performance. However, there is little theoretical framework or justification for using PPR. 
In section \ref{sec:ppr-gnn-pe}, we will propose a novel GNN framework and discuss the desired properties of proximity design, then justify the choice of PPR.


\subsection{Approx. PPR as $\ell_1$-Regularized Optimization}
The key ingredient in PPR-based GNNs is PPR calculation. Many algorithms have been designed to approximate PPR while having better run-time complexity.  
In this section, we define PPR and discuss the recent research on efficient PPR approximation, including traditional algorithmic approaches and newly emerging optimization perspectives.

\begin{definition}[Personalized PageRank] Given an undirected unweighted graph $\mathcal{G}(\mathcal{V}, \mathcal{E})$ with $|\mathcal{V}| = n $ and $ |\mathcal{E}| = m$. 
Define the random walk transition matrix $\bm P \triangleq \bm D^{-1} \bm A $, where $\bm A$ is the adjacency matrix of $\mathcal{G}$ and $\bm A = \bm A^{\top}$.
Given teleport probability $\alpha \in [0,1)$ and the source node $s$, the Personalized PageRank vector of $s$ is defined as:
\begin{equation}
\bm \pi_{\alpha, s} = (1-\alpha) \bm P^\top \bm \pi_{\alpha, s}  + \alpha \bm e_s, \label{equ:ppr}
\end{equation}
where the teleport probability $\alpha$ is a small constant (e.g. $\alpha=0.15$), $\bm e_s$ is an indicator column vector of node $s$, that is, $s$-th entry is 1, 0 otherwise. We use
$\bm \pi_{s, \alpha}$ \footnote{ We use column vector by default and may elide $s, \alpha$ for simplicity if there is no ambiguity in the context.} 
to denote the PPR of $s$ with teleport value $\alpha$.
\label{def:static-ppr-static-graph}
\end{definition}

\paragraph{Algorithmic approaches}: Power iteration and Forward push \cite{andersen2006local} (Algorithm \ref{algo:local-push} in Appendix)  are designed to compute PPR algorithmically. The key idea of the push-based methods \cite{wu2021unifying,wang2022edge} is to exploit PPR invariant\cite{andersen2006local}  and gradually reduce the residual vector to approximate the high-precision PPR.

\paragraph{Approximate PPR via optimization}: Interestingly, recent research \cite{fountoulakis2019variational, fountoulakis2022open} discovered that Algorithm \ref{algo:local-push} is equivalent to the Coordinate Descent algorithm, which solves an explicit optimization problem. 
Furthermore, with the specific termination condition, a $\ell_1$-regularized quadratic objective function has been proposed as the alternative PPR formulation (Lemma \ref{lemma:ppr-quad}).

\begin{lemma}[Variational Formulation of Personalized PageRank \cite{fountoulakis2019variational}] \label{lemma:ppr-quad}  Solving PPR in Equ.\ref{equ:ppr} is equivalent to solving the following $\ell_1$-regularized quadratic objective function where the PPR vector $\bm \pi := \bm D^{1/2} \bm x^{*} $: 

\begin{align}
     \bm x^{*} &= \argmin_{\bm x } f(\bm x) := \underbrace{\frac{1}{2} \bm x^{\top} \bm W \bm x + \bm x^{\top} \bm b}_{g(\bm x)}  + \underbrace{\epsilon\| \bm D^{1/2} \bm x \|_1}_{h(\bm x)}, \label{eq:ppr-quad-form} 
     \\where\   \bm W &= \bm D^{-1/2}(\bm D - (1-\alpha) \bm A ) \bm D^{-1/2} \nonumber , \ \bm b = - \alpha \bm D^{-1/2} \bm e_s   \nonumber
\end{align}

\end{lemma}

Lemma \ref{lemma:ppr-quad} provides a unique perspective to understand PPR and can be solved by proximal gradient methods such as ISTA.
%
Equ.\ref{eq:ppr-quad-form} has the composition of a smooth function  $g(\bm x)$
and a non-smooth function $h(\bm x)$, 
Note that 
\begin{align}
    \nabla^{2}g(\bm x) = \bm W &= \alpha \bm I_n+ (1-\alpha) \underbrace{(\bm I_n - \bm D^{-1/2} \bm A \bm D^{-1/2})}_{\text{Normalized Laplacian of $\lambda_{max} \leq 2$ }} ,
 \nonumber
\end{align}
where $\lambda_{max}$ is the largest eigenvalue. With $\| \nabla^2 g(\bm x) \|_2 \leq 2-\alpha$, let $\eta = \frac{1}{2-\alpha}$, one can convert the original problem (Equ.\ref{eq:ppr-quad-form}) into proximal form using the second-order approximation:
\begin{align}
    \bm x^{(k+1)} &= \textsc{prox}_{h(\cdot)}( \bm x^{(k)} - \eta  \nabla_i f(\bm x^{(k)})), where  \nonumber \\
    \textsc{prox}_{h(\cdot)}(\bm z) &= \argmin_x \frac{1}{2 \eta} \|\bm x - \bm z\|_2^2 + h(\bm x) \nonumber
\end{align}
Then, we can apply \textsc{ISTA} (Algorithm \ref{algo:ppr-ista}) to solve this well-defined PPR-equivalent $\ell_1$-regularized optimization problem. 




\vspace{-2mm}
\begin{algorithm}[ht]
\caption{$\textsc{ISTA Solver for PPR-equivalent Optimization}$ }
\begin{algorithmic}[1]
\State \textbf{Input: }$ k = 0, \bm x^{(0)} = \bm 0, \mathcal{G}, \epsilon, \alpha$
\State $\nabla f(\bm x^{(0)}) = - \alpha \bm D^{-1/2} \bm e_s$
\State $\eta = \frac{1}{2-\alpha}$

\While{$\bm x^{(k)}$ has not converged yet }
\State $ \bm z^{(k)}(i) := \bm x^{(k)}(i) - \eta  \nabla_i f(\bm x^{(k)}) $
\State $\bm x^{(k+1)(i)} = \textsc{prox}_{h(\cdot)}\left( \bm z^{(k)}(i)\right)$  can be solved analytically:
\[
    \bm x^{(k+1)}(i)= 
\begin{dcases}
    \bm z^{k}(i) -  \epsilon' \bm d(i)^{1/2} 
    , & \text{ if }  \bm z^{k}(i) >  \epsilon \bm d(i)^{1/2} \\
    0 , & \text{ if }   \| \bm z^{k}(i) \|_1 \leq  \epsilon' \bm d(i)^{1/2} \\
    \bm z^{k}(i) +  \epsilon' \bm d(i)^{1/2} , 
    & \text{ if }  \bm z^{k}(i) < - \epsilon \bm d(i)^{1/2} \\
\end{dcases}
\]
\EndWhile
\State \Return $\left( \bm x^{(k)}, \nabla f(\bm x^{(k)})\right)$
\State \textcolor{blue}{//Note that PPR vector $ \bm \pi = \bm D^{1/2} \bm x^{(k)}, \bm r =  \bm D^{1/2}\nabla f(\bm x^{(k)})$}

\end{algorithmic}
\label{algo:ppr-ista}
\end{algorithm}



\section{Problem Definition}
\label{sec:problem}

\begin{definition}[Node Classification Problem over Dynamic Graph]\label{def:prob-def}
Given an initial graph at timestamp $t=0$ $\mathcal{G}_0 = (\mathcal{V}, \mathcal{E}_0, \bm X_0)$
\footnote{For simplicity, we assume |$\mathcal{V}| = n$ does not change over time.},
 where $\bm X_0 \in \mathbb{R}^{n \times d}$ denotes the node attribute matrix, we want to predict the labels of a subset of nodes in $\mathcal{S} = \{u_0, u_1, \ldots, u_k\}$ as the graph evolves over time.
At time $t \in \{1, \ldots, T\}$, the graph $\mathcal{G}_{t-1}$ is updated with a batch of edge events $\Delta \mathcal{E}_{t}$ and a new node attribute $\bm X_t$. 
For each snapshot $\mathcal{G}_t$, we train a new classifier with updated node representation $\bm H_{t}^{\mathcal{S}} \in  \mathbb{R}^{|S| \times \bar d}$ to predict the node labels $\bm y_t$. 
\end{definition}


\paragraph{Key Challenges}: 
\xingzhiswallow{
Since \footnote{In practice, given a large graph we may use a small subset of nodes for training and evaluation.} 
$|\mathcal{S}| \ll |\mathcal{V}|$  the efficiency bottleneck for PPR-based models is at calculating $\bm H_{t}^{\mathcal{S}}$, rather than training prediction models given $\bm H_{t}^{\mathcal{S}}$ . 

}
 The key challenges of using PPR-based GNNs boil down to two questions -- \textbf{1):} \textit{How to efficiently calculate PPR as the graph evolves?} and \textbf{2):}  \textit{How to effectively fold in PPR as part of the node representation?} 
 In the following section, we answer these key questions using a justifiable PPR design with more robust and faster PPR-based GNNs over dynamic graphs. 





\section{Proposed Framework} 
\label{sec:method}
To efficiently and effectively learn node representations over time, we formulate PPR-based GNNs as a contextual node representation learning framework which has two key components: 
\textbf{1)}: Efficient PPR maintenance through an optimization lens; 
\textbf{2)}: Effective PPR-based node contextualization and positional encodings. 
Meanwhile, we discuss the desired properties of GNN design and justify the usage of PPR.
Then we instantiate a simple-yet-effective model called \textit{GoPPE}, analyze the overall complexity, and discuss its relations to other successful PPR-based GNN designs.


\subsection{Maintain PPR in the lens of Optimization}
Over an evolving graph, the new interactions between vertices are captured by adding edges, causing graph structure changes and making the previously calculated PPRs stale.  
To make the PPR-based GNNs reflect the latest graph structure, we propose to use ISTA as the PPR solver through an optimization lens and further explore its efficiency for dynamic graphs in various change patterns.

\paragraph{Naive Approach: Always from Scratch}:  One can simply re-calculate PPRs from scratch in each snapshot, however, this is sub-optimal because it does not leverage any previous PPR results, turning the dynamic graph into independent static graph snapshots. 
 


\paragraph{Improved Approach: Maintain PPR with warm starting in its optimization lens}: 
By exploiting the PPR invariant, PPR adjustment rules \cite{zhang2016approximate} have been proposed to maintain PPR w.r.t. a sequence of edge changes (add/delete edges), which can be proved equivalently to adjust the solution and gradient in the PPR's optimization formulation.
By adjusting the previous PPR and using it as the warm-starting point, we can maintain PPR incrementally. 
It provides an elegant solution to dynamically maintain PPR while enjoying a favorable convergence rate inherent in ISTA.

\begin{theorem}[PPR Adjustment Rules for Dynamic Graphs] Given a new edge inserted denoted as $(u, v, 1)$, 
The internal state of the PPR solver can be adjusted by the following rules to reflect the latest PPR but in compromised quality.
Instead of starting from $\bm x = \bm 0$, we use the adjusted solution as a warm-start for ISTA-solver (Algorithm. \ref{algo:ppr-ista}). 
\begin{align}
    \bm x'(u) &= \bm x(u) * \frac{\bm d(u)+1}{\bm d(u)} \label{equ:update-pi}\\
    \nabla f'(\bm x')(u) &= \nabla f(\bm x)(u) - \frac{\bm x(u)}{ \alpha \bm d^{1/2}(u)} \nonumber \\
    \nabla f'(\bm x')(v) &= \nabla f(\bm x)(v)  + \frac{(1-\alpha) }{\alpha} \frac{ \bm d^{-1/2}(v)*\bm x(u)}{ \bm d^{1/2}(u)} \nonumber , \\
\text{where }\bm x'(u), & \nabla f'(\bm x')\text{ denotes the value after adjustment.}  \nonumber 
\end{align}
\label{theorem:update}
\end{theorem}
Note that  $\nabla f'(x')$ can be evaluated analytically, so the above rules can be reduced to a single update using Equ.\ref{equ:update-pi}. Each update is a local operation having complexity $\mathcal{O}(1)$ per edge event for each node in $\mathcal{S}$.

\xingzhiswallow{
\begin{remark}
As defined in Appendix \ref{sec:proof-push-opt},  $\nabla f'(x')$ can be evaluated analytically, so the above rules can be reduced to a single update using Equ.\ref{equ:update-pi}. Each update is a local operation having complexity $\mathcal{O}(1)$ per edge event for each node in $\mathcal{S}$. 

\end{remark}

}


\xingzhiswallow{
    \begin{algorithm}[ht]
    \caption{$\textsc{ISTA PPR Solver for Dynamic Graphs}$ }
    \begin{algorithmic}[1]
    \State \textbf{Input: }$ k = 0, \bm x^{(0)} = \bm 0, \mathcal{G}, \epsilon, \alpha$
    \State $\nabla f(\bm x^{(0)}) = - \alpha \bm D^{-1/2} \bm e_s$
    \State $\eta = \frac{1}{2-\alpha}$
    
    
    \While{Termination Condition }
    \State $ \bm z^{(k)}(i) := \bm x^{(k)}(i) - \eta  \nabla_i f(\bm x^{(k)}) $
    \State $\bm x^{(k+1)(i)} = \textsc{prox}\left( \bm z^{(k)}(i)\right)$, which is solved analytically:
    \[
        \bm x^{(k+1)}(i)= 
    \begin{dcases}
        \bm z^{k}(i) -  \epsilon' \bm d(i)^{1/2} 
        , & \text{ if }  \bm z^{k}(i) >  \epsilon' \bm d(i)^{1/2} \\
        0 , & \text{ if }   \| \bm z^{k}(i) \|_1 \leq  \epsilon' \bm d(i)^{1/2} \\
        \bm z^{k}(i) +  \epsilon' \bm d(i)^{1/2} , 
        & \text{ if }  \bm z^{k}(i) < - \epsilon' \bm d(i)^{1/2} \\
    \end{dcases}
    \]
    \EndWhile
    
    \State \Return $\left(\bm \pi = \bm D^{1/2} \bm x^{(k)}, \bm r =  \bm D^{1/2}\nabla f(\bm x^{(k)}) \right)$
    \end{algorithmic}
    \label{algo:dyn-ista}
    \end{algorithm}

}

\subsection{PPR-based Contextual Node Representation and Positional Encodings}
\label{sec:ppr-gnn-pe}
Local message passing is the key ingredient in many GNN designs. Inspired by SGC, we provide another interpretation of GNN as a global node contextualization process. 
It aggregates node features in a global manner, in contrast to the viewpoint of recursive local neighbor message passing.   
We start the analysis from the simplified local message passing design. Recall the serialized GNNs in Equ. \ref{eq:sgc}:
\begin{align}
    \bm H ^{(L)} = \hat{\bm  W} \bm X \hat{\bm P}^{\top},\ 
    \hat{\bm  W} := \Pi_{l=0}^{L-1} \bm W^{(l)}, \ 
    \hat{\bm P} := (\bm D^{-1} \bm A)^{L}, \nonumber
\end{align}
where $L$ is the hyperparameter controlling how many hops should the GNN recursively aggregates the local neighbors' features. 
We can further organize Equ.\ref{eq:sgc} into $\bm H ^{(L)} = \hat{\bm  W}  \hat{\bm H}$, where $\hat{\bm H} = \bm X \hat{\bm P}^{\top}$. 
Let $\hat{\bm h_i}, \hat{\bm p_i}$ denote the $i^{th}$ column of $\hat{\bm H}$ and $\hat{\bm P}^{\top}$. $\hat{\bm p_i}(j)$ refers to its $j^{th}$ element. 
We re-organize Equ.\ref{eq:sgc} into vector form:
\begin{align}
    \hat{\bm h_i} &= \bm X \hat{\bm p_i} = \sum_{j \in \mathcal{V} } \hat{\bm p_i}(j) \bm x_j =  \sum_{j \in \mathcal{V} } \frac{ \hat{\bm p_i}(j)}{ z_i} \bm x_j, \label{equ:global-agg}\\
        where \ z_i &= \sum_{k \in \mathcal{V}} \hat{\bm p_i}(k) = \| \hat{\bm p_i}\|_1 = 1 \text{ and } \bm p_i(k) \geq 0 \label{equ:global-normalizer}
\end{align}

Regardless of the exact form of $\hat{\bm p_i}$, Equ.\ref{equ:global-agg} shows that $\hat{\bm h_i}$ can be decomposed into a weighted-sum formula with a proper normalizer $z_i$, perhaps thought of as \textit{"global attention over other nodes"} in deep learning terms. In this specific case, the attention weight $\hat{\bm p_i}$ is in its simplest form without any learnable parameter but it guarantees\footnote{$(\bm D^{-1} \bm A)^{k}$ guarantees to have row-wise $\ell_1$-norm equals to 1, and same for the column of its transpose. } that $\| \hat{\bm p_i}\|_1 = 1$ for any $L>0$ so that we do not need explicitly evaluate $z_i$ in a potentially expensive $|\mathcal{V}|$-length loop when $|\mathcal{V}|$ is large.
This observation provides an insightful viewpoint to interpret and design GNNs w.r.t. the node-wise attention weight $\hat{\bm p_i}(j)$. 
In the following, we summarize the desired properties of $\hat{\bm p_i}$ and justify our choice of using PPR (letting $\hat{\bm p_i} = \bm \pi_i$), then enhance the feature expressivity with PPR-based positional encodings.

\begin{definition}   
(The desired properties of node-wise attention design\label{def:att-property}) Given the formulation in Equ.\ref{equ:global-agg}, a well-designed node-wise attention vector $\hat{\bm p_i}$ should have the following properties:

\begin{itemize}
    \item Locality: Under the assumption of homophily graphs, $\hat{\bm p_i}(j)$ should capture the \textit{locality/proximity}  between node $u_i$ and $u_j$, making the feature aggregation informative. For example, the self-weight $\hat{\bm p_i}(i)$ should be emphasized.
    \item Efficiency: $\hat{\bm p_i}$ should be computed and maintained quickly, in particular, independently of node size $|\mathcal{V}|$.
    \item Sparsity: $\hat{\bm p_i}$  should be sparse $|\supp(\hat{\bm p_i})| \ll |\mathcal{V}|$ so that feature aggregation is memory bounded by $|\supp(\hat{\bm p_i})|$.
    \item Well-bounded normalizer $\| \hat{\bm p_i} \|_p$: The $\ell_p$-norm of $\hat{\bm p_i}$ should be bounded to avoid exhaustively calculating the normalizer $z_i$ over $\mathcal{V}$ when  $|\mathcal{V}|$ is large.
\end{itemize}

\end{definition}

\paragraph{A bad naive design}: One widely-adopted attention design is  $\hat{\bm p_i}(j):= \operatorname{CosSim}(\bm W \bm x_i, \bm W \bm x_j)$ where $\bm W$ is the learnt feature transformation. However, the resulting $\hat{\bm p_i}$ is neither sparse nor bounded and involves the node-wise calculation of complexity $\mathcal{O}(|\mathcal{V}|)$ per node, not to mention the ignorance of the graph structure.

\paragraph{PPR-based Attention Design and Justifications}:
To fulfill the desired properties, we use PPR vector as the attention design ($\hat{\bm p_i}:=\bm \pi_i$) with the following justifications:
1). PPR vector is a well-designed proximity measure over the graph with locality controlled by the teleport parameter $\alpha$, specifically, emphasizing the significance of node $u_i$ with the personalized vector $\bm e_i$.
2). PPR vector can be maintained efficiently with complexity $\mathcal{O}(\frac{1}{\epsilon \alpha})$ and independent of $|\mathcal{V}|$.
3). PPR vector is inherently sparse, explicitly revealed by the $\ell_1$-regularization in Equ.\ref{eq:ppr-quad-form} through optimization lens.
4). The $\ell_1$-norm of the PPR vector is  guaranteed to be 1, making the normalizer $z_i$ a constant and free of computation.  
Finally, we have a very simple-yet-effective global contextual node formulation :  
\begin{align}
    \hat{\bm h_i} = \bm X \hat{\bm \pi_i}
\end{align}

Although there are other designs for node-wise attention, 
To the best knowledge, we are the first to propose this framework with the desired properties and justify the choice of using PPR. 
This novel framework provides insights for future GNN design and motivates us to propose the following PPR-based positional encodings.

\paragraph{Robust Node Positional Encodings for Enhanced Locality Feature}:
The concept of \textit{"locality"} is crucial in many GNN designs as well as PPR formulation. 
In the original GCN design, the neighbor features are aggregated by $\bm D^{-1} (\bm I + \bm A)$ with the added self-loop to enhance its own features.  
On the other hand, PPR has the restart probability $\alpha \bm e_s$ to govern the locality.
%
%
Given the importance of the node locality information and the global contextualization framework, we seek to design a node Positional Encodings (PE) $\bm p_i \in \mathbb{R}^{d_{pe}}$ for node $u_i$. It is analogous to the token PE \cite{vaswani2017attention}  in NLP community to distinguish the tokens in different positions whereas node PE tells vertex's \textit{position} or \textit{address} in the graph.

We present the two simplest formulations of PE \footnote{We elide $\hat{\bm h_i}$ for simplicity; $\oplus$ is the element-wise addition operator. $||$ is the vector concatenation operator.} in Equ.\ref{equ:goppe} and then discuss its desired properties. Note that the node positional encodings are fused into the node $u_i$ to enhance its own locality feature, providing an attribute-agnostic representation.

\begin{align}
    \bm h_i =   
    \begin{dcases}
        \bm X \bm \pi_i \oplus \bm W_{pe} \bm p_i , \text{Additive PE} \\
        \bm X \bm \pi_i \  || \ \bm W_{pe} \bm p_i , \text{Concatenative PE}\\
    \end{dcases} \label{equ:goppe} 
\end{align}
where $\bm W_{pe} \in \mathbb{R}^{d \times d_{pe}}$ is the learnable mapping parameter to align dimensions with attributes for fusion. 

\begin{definition}[The Desired Properties of Node Positional Encodings \label{def:pos-encode}] Given the graph $\mathcal{G}$,  the positional encodings $\bm p_i$ for node $u_i$ is well-designed if it has the following properties:
\begin{itemize}
    \item Effectiveness: $\bm p_i$ should represent the node's \textit{"position"} or \textit{"locality"} over the graph and be distinguishable among vertices even if the node attributes are indistinguishable. 
    For example, $\bm p_i \neq \bm p_j \text{ while } \bm x_i = \bm x_j, \forall i,j \in \mathcal{V}$.
    \item Efficiency:  $\bm p_i$ should be computationally cheap $\ll \mathcal{O}(|\mathcal{V}|)$ and low-dimensional with $d_{pe} \ll |\mathcal{V}|$ if $|\mathcal{V}|$ is large.
\end{itemize}
\end{definition}

\paragraph{PPR-based node positional encodings and justifications}: 
Interestingly, as the by-product of the PPR-based node contextualization, we re-use the unique and sparse PPR vector $\bm \pi_i \in \mathbb{R}^{|\mathcal{V}|}, \forall i \in \mathcal{S}$ as the source to derive the node positional encoding $\bm p_i \in \mathbb{R}^{d_{pe}}$ using  simple dimension reduction operations. 
Specifically, we can use sparse random projection 
or hashing-based dimension reduction (Line \ref{proc:HashReduce} in Algorithm \ref{algo:goppe}).
The reduction operation has time complexity $\mathcal{O}(|\supp(\bm \pi_i)|) \ll\mathcal{O}( |\mathcal{V}|)$ per node, and maximizes the usage of PPR vector. 
Meanwhile, the resulting $\bm p_i$ encodes a strong vertex positional signal per se. In an extreme case, where the node attributes are weak (noisy), $\bm p_i$ alone can provide good-quality embeddings from pure network structure, consistently providing a robust node representation.


\paragraph{GoPPE as an instantiated model}: Putting all together, we instantiate a very simple-yet-effective GNN called \textit{GoPPE} for dynamic graphs. Although there are more sophisticated neural network designs under the proposed framework, we aim to demonstrate its usefulness with the simplest choices. We present the detailed model components in Algorithm \ref{algo:goppe}.

\paragraph{Relations to other PPR-based GNNs}:
Interestingly, many successful PPR-based GNNs 
\footnote{we assume that all use the standard PPR formulation with transitional matrix $\bm D^{-1} \bm A$} 
fit the proposed framework with various design choices. 
For example, \textsc{DynamicPPE} and \textsc{InstantEmb} can be considered as the feature-less \textsc{GoPPE} with only positional encodings, which is robust when the node attribute is noisy.  Meanwhile, \textsc{InstantGNN} is equivalent to \textsc{GoPPE} without positional encodings and chooses \textsc{ForwardPush} as the PPR solver. 
We summarize their design choices in Table \ref{tab:design-compare} within our proposed framework.


\paragraph{Complexity Analysis}:
Given total $T$ graph snapshots with $m$ edge events and  $|\mathcal{S}| = K$, the complexity of \textsc{GoPPE} ( Algorithm \ref{algo:goppe}) involves four parts:

\begin{itemize}
    \item 1: $\mathcal{O}(\frac{KT}{\epsilon \alpha})$ for the \textsc{ISTA} solver for PPR calculation 
    . 
    \item 2: $ \mathcal{O}(KT \operatorname{max}(|\supp(\pi_i)|), \forall i \in \mathcal{S}) $ for the positional encoding calculation hashing method. 
    \item 3: $ \mathcal{O}(KTm) $ for the PPR adjustment of $m$ edge events.
    \item 4: $ \mathcal{O}(KT) $ for neural network training/inference. Since we use the simple stacked MLPs, we consider roughly $ \mathcal{O}(1) $ per sample.
\end{itemize}
To sum up, the total complexity is $ \mathcal{O}(K T m + \frac{K T}{\epsilon \alpha}) $. Note that the complexity is independent of $|\mathcal{V}|$, indicating the algorithm only involves efficient local computation and is scalable to large graphs.




\xingzhiswallow{
    
    The overall complexity of tracking subset of $k$ nodes across $T$ snapshots depends on run time of three main steps as shown in Algo. \ref{algo:dynanom-node}: 1. the run time of \textsc{IncrementPush} for nodes $\mathcal{S}$; 2. the calculation of dynamic node representations, which be finished in $\mathcal{O}(k T |\supp(\bm p_{s^\prime})|)$ where $|\supp(\bm p_{s^\prime})|$ is the maximal support of all $k$ sparse vectors, i.e. $|\supp(\bm p_{s^\prime})| = \max_{s\in \mathcal{S}} |\supp(\bm p_{s})|$. The overall time complexity of our proposed framework is stated as in the following theorem.
    [Time Complexity of \textsc{DynAnom}]
    Given the set of edge events where $\mathbb{E} = \{ e_1, e_2,\ldots, e_m\}$ and $T$ snapshots and subset of target nodes $\mathcal{S}$ with $|\mathcal{S}| = k$, the instantiation of our proposed \textsc{DynAnom} detects whether there exist events in these $T$ snapshots in $\mathcal{O}( k m/\alpha^2 + k \bar{d^t} / (\epsilon \alpha^2) +  k/(\epsilon \alpha) + k T s)$ where $\bar{d^t}$ is the average node degree of current graph $\mathcal{G}_t$ and $s\triangleq \max_{s\in\mathcal{S}} |\supp(\bm p_s)|$.
}


\section{Experiments}
\label{sec:exp}
In this section, we describe the experiment configurations and present and discuss the results of running time and prediction accuracy.  
Finally, we demonstrate the effectiveness of the introduced positional encodings when the node attributes are noisy. 

\subsection{Experiment Configurations}
\label{sec:dataset}
\paragraph{Datasets:}
To evaluate GNNs in a dynamic setting, we divide the edges of static undirected graphs into snapshots \footnote{
Although such conversion breaks the temporal correlation, we focus on a more fundamental problem of node embedding updates instead of temporal relations, which is similar to \cite{zheng2022instant}.
}.   
For node classification tasks, the first graph snapshot $\mathcal{G}_0$ contains the $50\%$ of the total edges, and the other half is evenly divided into edge events in each snapshot $\Delta \mathcal{E}_{i}, i \in \{1, \ldots, T-1\}$ for the following $\mathcal{G}_{ \{1, \ldots, T-1\} }$. 
We fix the node labels in each snapshot, while the node attributes may change according to the experiment settings.  All the train/dev/test nodes (70\%/10\%/20\%) are sampled from the first snapshots without any dangling node. 
In the efficiency test, we report the running time in two dynamic setting detailed in Section \ref{sec:ppr-speed}. 
The statistics of datasets are presented in Table \ref{tab:dataset}.

\xingzhiswallow{

    \begin{table}[htbp]
        \centering
            \caption{Datasets used in our experiments. $|\mathcal{V}|$: number of nodes. $|\mathcal{E}|$ number of edges ,  $T$: number of snapshots for node classification, $|\mathcal{S}|$: total nodes in train/test sets, $d$: dimension of node raw feature, $|\mathcal{L}|$: number of node labels. }
        \label{tab:dataset}
        \vspace{-3mm}
    
        \footnotesize
        \begin{tabular}{
        p{0.06\textwidth} 
        |p{0.03\textwidth}
        R{0.05\textwidth} 
        R{0.06\textwidth}
        R{0.01\textwidth}
        R{0.03\textwidth}
        R{0.03\textwidth}
        R{0.035\textwidth}}
        \toprule
        Dataset &   Size &  $|\mathcal{V}|$ &  $|\mathcal{E}|$ &  $T$ & $|\mathcal{S}|$  &  $d$ &  $|\mathcal{L}|$ \\
        \midrule
            Cora &  small &           2,708 &                    5,277 &              5 &   1,000 &         1,433 &                    7 \\
        Citeseer &  small &           3,279 &                    4,551 &              5 &   1,000 &         3,703 &                    6 \\
          Pubmed &  small &          19,717 &                   44,323 &              5 &   1,000 &           500 &                    3 \\ 
          Flickr & medium &          89,250 &                  449,877 &              5 &   1,000 &           500 &                    7 \\
      Arvix & medium &         169,343 &                1,157,798 &              5 &   1,000 &           128 &                   40 \\ 
     BlogCatalog &  small &           5,196 &                  171,742 &              5 &   1,000 &         8,189 &                    6 \\
        DBLP &  small &          17,716 &                   52,866 &              5 &   1,000 &         1,639 &                    4 \\
          Physics    &  small &          34,493 &                  247,961 &              5 &   1,000 &         8,415 &                    5 \\
        \bottomrule
        \end{tabular}
    \end{table}
}

\begin{table}[htbp]
        \centering
            \caption{Datasets used in our experiments. $|\mathcal{V}|$: number of nodes. $|\mathcal{E}|$ number of edges ,  $T$: number of snapshots for node classification, $|\mathcal{S}|$: total nodes in train/test sets, $d$: dimension of node raw feature, $|\mathcal{L}|$: number of node labels. }
        \label{tab:dataset}
    
        \footnotesize
        \begin{tabular}{
        p{0.08\textwidth} 
        | R{0.05\textwidth} 
        R{0.06\textwidth}
        R{0.01\textwidth}
        R{0.03\textwidth}
        R{0.03\textwidth}
        R{0.035\textwidth}}
        \toprule
        Dataset &    $|\mathcal{V}|$ &  $|\mathcal{E}|$ &  $T$ & $|\mathcal{S}|$  &  $d$ &  $|\mathcal{L}|$ \\
        \midrule
            Cora &             2,708 &                    5,277 &              5 &   1,000 &         1,433 &                    7 \\
        Citeseer &             3,279 &                    4,551 &              5 &   1,000 &         3,703 &                    6 \\
     BlogCatalog &            5,196 &                  171,742 &              5 &   1,000 &         8,189 &                    6 \\
        DBLP &            17,716 &                   52,866 &              5 &   1,000 &         1,639 &                    4 \\
          Pubmed &            19,717 &                   44,323 &              5 &   1,000 &           500 &                    3 \\ 
    Physics    &           34,493 &                  247,961 &              5 &   1,000 &         8,415 &                    5 \\
          Flickr &          89,250 &                  449,877 &              5 &   1,000 &           500 &                    7 \\
      Arvix &          169,343 &                1,157,798 &              5 &   1,000 &           128 &                   40 \\ 
        \bottomrule
        \end{tabular}
    \end{table}


\xingzhiswallow{
    We apply the stratified sampling and add at least one sample of each class into the train/dev/test set. The total sampled nodes are presented $|\mathcal{S}|$ in Table. \ref{tab:dataset}.
    }


\paragraph{Baselines:} Besides the MLPs and GCN as the  baselines, we compare with the State-of-the-Art PPR-based methods as listed in Table \ref{tab:design-compare}.
Within our framework, we implement the baselines which are equivalent to \textsc{DynamicPPE}, \textsc{PPRGo} and \textsc{InstantGNN}, together with our proposed \textsc{GoPPE}.\footnote{We denote \textsc{DynamicPPE} as \textsc{DynPPE}, \textsc{InstantGNN} as \textsc{InsGNN}.}  We apply the same shallow neural classifier (2-layer MLPs) and hyper-parameter settings to test the embedding quality across all methods for a fair comparison. More details are presented in the Appendix \ref{app:train-detail}.

\paragraph{Metrics:} 
Since the PPR weights calculating are the most time-consuming bottleneck, we measure the total CPU time for PPR calculation along with the graph updates. 
We use the multi-class classification setting, train the classifier separately for each snapshot
\footnote{For benchmarking purposes, we train the classifier individually for each snapshot to minimize the variability caused by continual learning.}
, then report the testing accuracy. 
We repeat all experiments three times with various random seeds and report the average value.



\xingzhiswallow{
    \begin{itemize}
        \item \textcolor{red}{Below: Feature-based}
        \item baseline: MLPs
        \item baseline: DynPPE
        \item baseline: \sout{GCN}
        \item baseline: \sout{GRPGNN}
        \item baseline: PPRGO (agg top-k)
        \item baseline: InstantGNN (mixrank-all: agg all, original code is not released)
        \item \textcolor{red}{Below: our proposed baselines:}
        \item baseline: MixRank-rnd (agg rand-k)
        \item baseline: GoPPE (MixRank-all + DynPPE)
        \item baseline: \sout{MixRank-percentile (agg top-10\%,20\%, ... )}
        \item baseline: \sout{ MixRank-poly (fit a polynomial to modify ppr as agg weights. )}
\end{itemize}
}

\xingzhiswallow{
\begin{tabular}{llrrrr}
\toprule
ppr-algo &  increment &   cora &  citeseer &  pubmed &   flickr \\
\midrule
    push &      False & 288.07 &    149.89 & 3254.14 & 19035.70 \\
    ista &      False & 150.53 &    148.38 & 1376.25 &  7452.47 \\
    push &       True & 430.57 &    223.41 & 4786.90 & 31889.43 \\
    ista &       True & 137.21 &    121.29 & 1212.33 &  7249.87 \\
\bottomrule
\end{tabular}

\begin{tabular}{llrrrr}
\toprule
ppr-algo &  increment &     cora &  citeseer &   pubmed &   flickr \\
\midrule
    push &      False & 3.99e-08 &  2.25e-08 & 4.57e-08 & 4.32e-08 \\
    ista &      False & 5.22e-08 &  2.78e-08 & 6.03e-08 & 2.84e-08 \\
    push &       True & 4.06e-08 &  2.30e-08 & 4.65e-08 & 4.34e-08 \\
    ista &       True & 5.14e-08 &  2.73e-08 & 5.94e-08 & 2.80e-08 \\
\bottomrule
\end{tabular}

}


\subsection{Efficiency Experiments } 
\label{sec:ppr-speed}
In PPR-based GNNs for dynamic graphs, the main computational bottleneck is the PPR maintenance over time. 
In this section, we present the running time under two cases regarding the intensity of graph changes between snapshots 
-- \textbf{1: Major change case} where we start from $50\%$ of the total edges, then add $10\%$ edges between snapshots until all edges are used. 
\footnote{Same as we described in Section \ref{sec:dataset}, we also evaluate  accuracy in this case.} 
\textbf{2: Minor change case} where we start from  a almost full graph but held out a few edges, then we add a fixed number of edges (e.g., 100) between snapshots until all edges are added. We design these two cases to evaluate the efficiency of PPR maintenance under different dynamic patterns. In the following, we discuss the insights from our empirical observations.

\begin{table}[ht]
    \centering
    \caption{The total CPU time (in seconds) for PPR maintenance in major change case. 
    \text{GoPPE} uses the least time to obtain the approximate PPR weights with similar $\ell_1$-error. We elide \textsc{DynPPE} since it use the same PPR solve as used in \textsc{InsGNN}.
    }
    \vspace{-2mm}
    \label{tab:cpu-time}
\small

\xingzhiswallow{
\begin{tabular}{lrrrrrrrr}
\toprule
Method &  Cora & Citeseer &   Pubmed & Arxiv &    Flickr & BlogCatalog &   DBLP &  Physics \\
\midrule
    PPRGo  & 288.07 &   149.89 & 3,254.14 &  79,133.65 & 19,035.70 &    3,656.59 & 631.02 & 2,941.66 \\
    InsGNN  & 430.57 &   223.41 & 4,786.90 & 108,048.49 & 31,889.43 &    5,436.69 & 849.92 & 4,571.53 \\
    GoPPE-S  & 150.53 &   148.38 & 1,376.25 &  45,320.18 &  7,452.47 &    2,391.94 & 291.98 & 1,247.90 \\
    \textbf{GoPPE-D} & \textbf{137.21} &   \textbf{121.29} & \textbf{1,212.33} &  \textbf{38,830.99} &  \textbf{7,249.87} &    \textbf{2,243.40} & \textbf{264.11} & \textbf{1,169.70} \\
\bottomrule
\end{tabular}

}

\begin{tabular}{lrrrr}
\toprule
    & PPRGo &       InsGNN &      GoPPE-S &      \textbf{GoPPE-D} \\
\midrule
Cora &   288.07 &     430.57 &    150.53 &    \textbf{137.21} \\
Citeseer &   149.89 &     223.41 &    148.38 &    \textbf{121.29} \\
Pubmed & 3,254.14 &   4,786.90 &  1,376.25 &  \textbf{1,212.33} \\
Arxiv &79,133.65 & 108,048.49 & 45,320.18 & \textbf{38,830.99} \\
Flickr &19,035.70 &  31,889.43 &  7,452.47 &  \textbf{7,249.87} \\
BlogCatalog & 3,656.59 &   5,436.69 &  2,391.94 &  \textbf{2,243.40} \\
DBLP &   631.02 &     849.92 &    291.98 &    \textbf{264.11} \\
Physics & 2,941.66 &   4,571.53 &  1,247.90 &  \textbf{1,169.70} \\
\bottomrule
\end{tabular}
\end{table}

\paragraph{GoPPE is efficient and scalable in major change case}: Table \ref{tab:cpu-time} presents the total running time of PPR calculation. Our proposed \textsc{GoPPE} is consistently faster than others while maintaining a similar PPR precision. 
In general, \textsc{GoPPE} can achieve upto 6 times faster than forward push-based methods. The gain is more significant over large graphs. 
The advantage mainly comes from the ISTA-based PPR solver, which enjoys both the ISTA's favorable convergence rate and the warm-starting strategy as the incremental PPR maintenance method. 
Note that even the static \textsc{GoPPE-S} is faster than the incremental Push-based method. 

\begin{table}[ht]
\centering
    \caption{The total CPU time (in seconds) used in the minor change case.  The incremental PPR maintenance methods (\textsc{GoPPE-D} and \textsc{InsGNN}) achieve significantly better efficiency than their static counterparts. In addition, \textsc{GoPPE-D} is steadily faster, and even the static \textsc{GoPPE-S} is comparable to the incremental push-based methods.  Note that the reported CPU time is not directly comparable to Table \ref{tab:cpu-time} due to different experimental settings. }
    \label{tab:cpu-time-small-change}
\small
\xingzhiswallow{
\begin{tabular}{lrrrrrrrr}
\toprule
     Method &  Cora & Citeseer &   Pubmed & Arxiv &    Flickr & BlogCatalog &   DBLP & Physics \\ \midrule
    PPRGo & 548.24 &   135.96 & 5,852.38 & 138,513.10 & 39,692.24 &   10,454.92 & 1,643.46 & 8,290.49 \\
    InsGNN & 358.51 &    87.17 & 2,896.12 &  24,742.79 & 11,883.08 &    8,852.43 &   770.11 & 5,311.04 \\
    GoPPE-S & 319.60 &   142.20 & 2,847.84 &  75,117.79 & 15,632.07 &    6,941.65 &   812.63 & 3,492.90 \\
    \textbf{GoPPE-D} & \textbf{190.16} &    \textbf{64.77} & \textbf{1,495.70} &  \textbf{22,512.81 }&  \textbf{7,462.86} &    \textbf{4,137.82} &   \textbf{419.52} & \textbf{1,885.49}\\
\bottomrule
\end{tabular}
}

\begin{tabular}{lrrrr}
\toprule
    & PPRGo &       InsGNN &      GoPPE-S &      \textbf{GoPPE-D} \\
\midrule
Cora &    548.24 &    358.51 &    319.60 &    \textbf{190.16 }\\
Citeseer &    135.96 &     87.17 &    142.20 &     \textbf{64.77} \\
Pubmed &  5,852.38 &  2,896.12 &  2,847.84 &  \textbf{1,495.70} \\
Arxiv &138,513.10 & 24,742.79 & 75,117.79 & \textbf{22,512.81} \\
Flickr & 39,692.24 & 11,883.08 & 15,632.07 &  \textbf{7,462.86} \\
BlogCatalog & 10,454.92 &  8,852.43 &  6,941.65 &  \textbf{4,137.82} \\
DBLP &  1,643.46 &    770.11 &    812.63 &    \textbf{419.52} \\
Physics &  8,290.49 &  5,311.04 &  3,492.90 &  \textbf{1,885.49} \\
\bottomrule
\end{tabular}

\end{table}


\xingzhiswallow{

\begin{table}[ht]
    \centering
    \caption{The total CPU time (in seconds) for PPR maintenance in major change case. 
    \text{GoPPE} uses the least time to obtain the approximate PPR weights with similar $\ell_1$-error. We elide \textsc{DynPPE} since it use the same PPR solve as used in \textsc{InsGNN}.
    }
    \vspace{-2mm}
    \label{tab:cpu-time}
    \begin{tabular}{lrrrrr}
    \toprule
    Method &     cora &  citeseer &  pubmed &  arxiv &   flickr \\
    \midrule

        \textsc{PPRGo} & 288.07 &    149.89 & 3,254.14 &    79,133.65 & 19,035.70 \\  \hline
        \textsc{InsGNN} &   430.57 &    223.41 & 4,786.90 &  108,048.49 & 31,889.43 \\ \hline
        \textsc{GoPPE-S}  & 150.53 &    148.38 & 1,376.25 &    45,320.18 &  7,452.47\\ \hline
        \textbf{\textsc{GoPPE-D}} &  \textbf{137.21} &    \textbf{121.29} &  \textbf{1,212.33} &    \textbf{38,830.99} &  \textbf{7,249.87} \\

    \bottomrule
    \end{tabular}
\end{table}

}

\begin{remark}
    \textsc{ForwardPush} and \textsc{ISTA} has the same theoretical worst-case complexity $\mathcal{O}(\frac{1}{\alpha \epsilon})$, the observed running time difference is caused by their formulations where algorithmic \textsc{ForwardPush} is implemented by queue, while  \textsc{ISTA} adopts numerical computing on cache-friendly continuous memory. Similar observations are also in \cite{wu2021unifying}.
\end{remark}

\begin{figure}[ht]
    \centering
    \subfloat[Major changes case\label{fig:ppr-time-large-change}]{
        \includegraphics[width=0.48\linewidth]{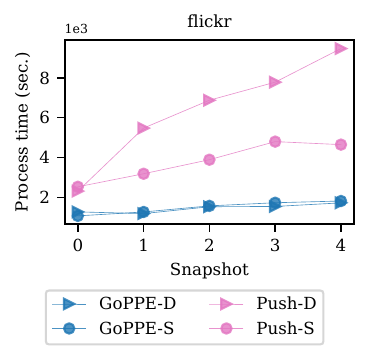}

    }
    \hfill
    \subfloat[Minor changes case \label{fig:ppr-time-small-change}]{
        \includegraphics[width=0.48\linewidth]{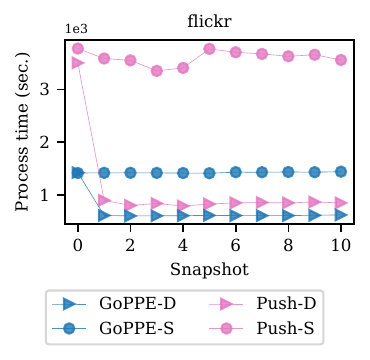}
    }
    \vspace{-3mm}
    \caption{Subfigures (a) and (b) show the CPU time  in \textit{Major/Minor} graph change experiment settings, respectively. The intensity of graph changes between snapshots (major vs minor) may greatly affect the PPR updating time. In both cases, \textsc{GoPPE} is consistently efficient in terms of total CPU time. \textsc{PPRGo} uses \textsc{Push-S} while \textsc{DynPPE} and \textsc{InsGNN} use  \textsc{Push-D}.   Note that the absolute time is not directly comparable between subfigures due to different settings.  }
    \label{fig:ppr-time-wrt-edge-changes}
\end{figure}

\paragraph{The efficiency of Push-based method is limited in major change case while boosts up in minor change case}: 
One interesting observation is that the incremental push-based method (\textit{Push-D} used by \textsc{DynPPE} and \textsc{InsGNN}  ) is not faster than their static counterparts (\textit{Push-S} used by \textsc{PPRGo}), as presented in Table \ref{tab:cpu-time} and Figure \ref{fig:ppr-time-large-change}.
The root reason relates to how \textit{"different"} the graph will be after the edge events. In major change case, we cause the graph changes drastically between snapshots by adding a large number of edges. 
Consequently, the quality of the approximate PPR is too compromised after applying the PPR update rules. 
Unfortunately, the update rules create negative residuals and incur more push operations than their static version \footnote{Static push method only has a positive and monotonic decreasing residual.}. When the graph changes drastically, it eventually exceeds the total operations required from scratch to maintain the PPR precision. 



\xingzhiswallow{

\begin{table}[ht]
    \centering
    \caption{The total CPU time (in seconds) used in the minor change case.  The incremental PPR maintenance methods (\textsc{GoPPE-D} and \textsc{InsGNN}) achieve significantly better efficiency than their static counterparts. In addition, \textsc{GoPPE-D} is steadily faster, and even the static \textsc{GoPPE-S} is comparable to the incremental push-based methods.  Note that the reported CPU time is not directly comparable to Table \ref{tab:cpu-time} due to different experimental settings. }
    \label{tab:cpu-time-small-change}
    \begin{tabular}{llrrrrr}
    \toprule
    Method & cora &  citeseer &  pubmed & arxiv&  flickr \\
    \midrule
       \textsc{PPRGo} &     548.24 &    135.96 & 5,852.38 & 138,513.10  & 39,692.24 \\ \hline
        \textsc{InsGNN} &   358.51 &    87.17 & 2,896.12 & 24,742.79  & 11,883.08 \\ \hline
       \textsc{GoPPE-S} &   319.60 &    142.20 & 2,847.84 & 75,117.79 & 15,632.07 \\ \hline
        \textbf{\textsc{GoPPE-D}} &  \textbf{190.16} &    \textbf{64.77} & \textbf{1,495.70} & \textbf{22,512.81} & \textbf{7,462.86} \\
    \bottomrule
    \end{tabular}
\end{table}

}

\paragraph{GoPPE is consistently faster regardless of the graph change intensity}: 
Table \ref{tab:cpu-time-small-change} and Figure \ref{fig:ppr-time-small-change} present PPR efficiency results in the minor change case where only a small number of edges are added between snapshots.
It shows that the update rules greatly boost PPR efficiency for both increment PPR maintenance methods (\textsc{Push-D} and \textsc{GoPPE-D}). 
Besides, \textsc{GoPPE} is a steadily fast method,  achieving impressive speed consistently in both cases.  
%









\xingzhiswallow{

    \subsection{Results on OGB-Arvix (Homophily Graph)}
    \begin{figure}[ht]
        \centering
        \subfloat[][accuracy]{
         \includegraphics[width=0.5\textwidth]{figs/exp-1-v7/ogbn-arxiv-debug-7/fig-clf-res.pdf}
        } 
        \\
        \subfloat[][ppr running time]{
         \includegraphics[width=0.5\textwidth]{figs/exp-1-v7/ogbn-arxiv-debug-7/fig-ppr-time-alpha-0.15.pdf}
        }
        \caption{ogbn-arxiv-res}
        \label{fig:clf-ogbn-arxiv}
    \end{figure}
    \newpage

    \subsection{Results on Cora (Homophily Graph)}
    \begin{figure}[ht]
        \centering
        \subfloat[][accuracy]{
    \includegraphics[width=0.5\textwidth]{figs/exp-1-v7/cora-debug-7/fig-clf-res.pdf}
        }
        
        \subfloat[][ppr running time]{
            \includegraphics[width=0.5\textwidth]{figs/exp-1-v7/cora-debug-7/fig-ppr-time-alpha-0.15.pdf}
        }
        \caption{Cora-res}
        \label{fig:clf-cora}
    \end{figure}
    \newpage

    \subsection{Results on Citeseer (Homophily Graph)}
    \begin{figure}[ht]
        \centering
        \subfloat[][accuracy]{
            \includegraphics[width=0.45\textwidth]{figs/exp-1-v7/citeseer-debug-7/fig-clf-res.pdf}
        }
    
        \subfloat[][ppr running time]{
            \includegraphics[width=0.5\textwidth]{figs/exp-1-v7/citeseer-debug-7/fig-ppr-time-alpha-0.15.pdf}
        }
        \caption{citeseer-res}
        \label{fig:clf-citeseer}
    \end{figure}
    \newpage

    \subsection{Results on PubMed (Homophily Graph)}
    \begin{figure}[ht]
        \centering
        \subfloat[][accuracy]{
            \includegraphics[width=0.45\textwidth]{figs/exp-1-v7/pubmed-debug-7/fig-clf-res.pdf}
        }
    
        \subfloat[][ppr running time]{
            \includegraphics[width=0.5\textwidth]{figs/exp-1-v7/pubmed-debug-7/fig-ppr-time-alpha-0.15.pdf}
        }
        \caption{pubmed-res}
        \label{fig:clf-pubmed}
    \end{figure}
    \newpage

    \subsection{Results on Flickr (Homophily Graph)}
    \begin{figure}[ht]
        \centering
            \subfloat[][accuracy]{
        \includegraphics[width=0.5\textwidth]{figs/exp-1-v7/flickr-debug-7/fig-clf-res.pdf}
        }
    
        \subfloat[][ppr running time]{
        \includegraphics[width=0.5\textwidth]{figs/exp-1-v7/flickr-debug-7/fig-ppr-time-alpha-0.3.pdf}
        }
    
        \caption{flickr-res}
        \label{fig:clf-flickr}
    \end{figure}
    \newpage

    \subsection{Results on Film (Heterophily Graph)}
    \begin{figure}[ht]
        \centering
        \includegraphics[width=0.5\textwidth]{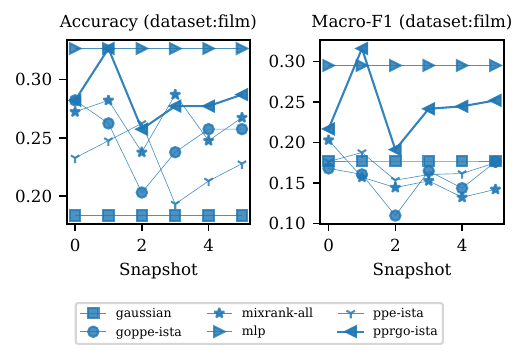}
        \caption{film-res}
        \label{fig:clf-film}
    \end{figure}
    \begin{table}[ht]
\centering
\caption{Avg. Accuracy (film)}
\label{tab:clf-res-film-avg-acc}
\begin{tabular}{lr}
\toprule
 & accuracy-mean \\
model &  \\
\midrule
mlp & 0.3267 \\
pprgo-ista & 0.2847 \\
mixrank-all & 0.2657 \\
goppe-ista & 0.2500 \\
ppe-ista & 0.2294 \\
gaussian & 0.1832 \\
\bottomrule
\end{tabular}
\end{table}

\begin{table}[ht]
\centering
\caption{Avg. Macro-F1 (film)}
\label{tab:clf-res-film-avg-macro-f1}
\begin{tabular}{lr}
\toprule
 & macro-f1-mean \\
model &  \\
\midrule
mlp & 0.2950 \\
pprgo-ista & 0.2437 \\
gaussian & 0.1772 \\
ppe-ista & 0.1692 \\
mixrank-all & 0.1554 \\
goppe-ista & 0.1539 \\
\bottomrule
\end{tabular}
\end{table}

    \newpage
    
    \subsection{Results on Chameleon (Heterophily Graph)}
    \begin{figure}[ht]
        \centering
        \includegraphics[width=0.5\textwidth]{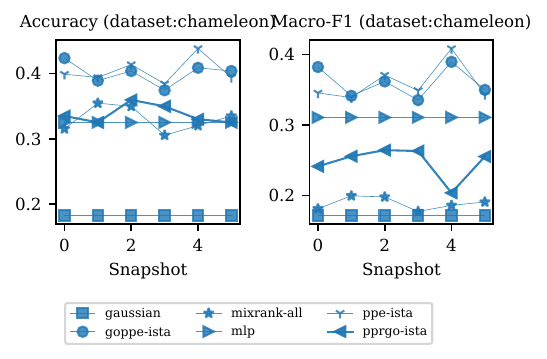}
        \caption{chameleon-res}
        \label{fig:clf-chameleon}
    \end{figure}
    \begin{table}[ht]
\centering
\caption{Avg. Accuracy (chameleon)}
\label{tab:clf-res-chameleon-avg-acc}
\begin{tabular}{lr}
\toprule
 & accuracy-mean \\
model &  \\
\midrule
ppe-ista & 0.4039 \\
goppe-ista & 0.4007 \\
pprgo-ista & 0.3374 \\
mixrank-all & 0.3300 \\
mlp & 0.3251 \\
gaussian & 0.1823 \\
\bottomrule
\end{tabular}
\end{table}

\begin{table}[ht]
\centering
\caption{Avg. Macro-F1 (chameleon)}
\label{tab:clf-res-chameleon-avg-macro-f1}
\begin{tabular}{lr}
\toprule
 & macro-f1-mean \\
model &  \\
\midrule
goppe-ista & 0.3599 \\
ppe-ista & 0.3592 \\
mlp & 0.3107 \\
pprgo-ista & 0.2472 \\
mixrank-all & 0.1888 \\
gaussian & 0.1716 \\
\bottomrule
\end{tabular}
\end{table}

    \newpage
    
    \subsection{Results on Squirrel (Heterophily Graph)}
    \begin{figure}[ht]
        \centering
        \includegraphics[width=0.5\textwidth]{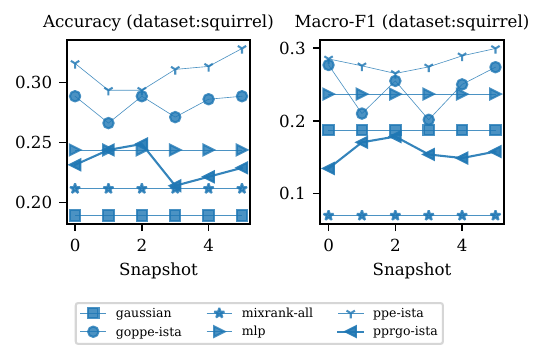}
        \caption{squirrel-res}
        \label{fig:clf-squirrel}
    \end{figure}
    \begin{table}[ht]
\centering
\caption{Avg. Accuracy (squirrel)}
\label{tab:clf-res-squirrel-avg-acc}
\begin{tabular}{lr}
\toprule
 & accuracy-mean \\
model &  \\
\midrule
ppe-ista & 0.3093 \\
goppe-ista & 0.2815 \\
mlp & 0.2438 \\
pprgo-ista & 0.2313 \\
mixrank-all & 0.2114 \\
gaussian & 0.1891 \\
\bottomrule
\end{tabular}
\end{table}

\begin{table}[ht]
\centering
\caption{Avg. Macro-F1 (squirrel)}
\label{tab:clf-res-squirrel-avg-macro-f1}
\begin{tabular}{lr}
\toprule
 & macro-f1-mean \\
model &  \\
\midrule
ppe-ista & 0.2815 \\
goppe-ista & 0.2445 \\
mlp & 0.2368 \\
gaussian & 0.1875 \\
pprgo-ista & 0.1573 \\
mixrank-all & 0.0698 \\
\bottomrule
\end{tabular}
\end{table}

    \newpage
    
}



\subsection{Accuracy and Robustness Experiments}
 We have two node classification settings described as follows to evaluate the performance and robustness of \textsc{GoPPE}. In both cases, we adopt \textit{major change} as the dynamic pattern.
 \textbf{1: Intact case} where we use the intact raw node attributes for global feature aggregation; \textbf{2: Noisy case} where we add Gaussian noise to the node attributes (similar to \cite{liu2021graph}) at the beginning, then gradually reduce the noise level to simulate the node attributes getting better over time.  
The added noise is monotonically reducing defined as: 
\begin{align}
    \hat{\bm  x_t} =  \lambda_t \bm x_t + (1-\lambda_t) \mathcal{N}(\mu, \sigma^2),  \label{equ:noisy}
\end{align}
where  $\lambda_t = \frac{t}{T} + \lambda_{base} \in [0, 1]$ controls noise level,  $t$ is the snapshot stamp, $T$ is the total number of snapshots, $\lambda_{base} \in [0,1]$ controls the initial noise level, $\mu, \sigma^2 $ are the mean and variance of the elements in node attributes matrix $\bm X$, respectively. As $t$ increases from $0$ to $T$, the noisy component decays linearly. 
We use this noise model to simulate the feature refinement process, similar to the volunteers helping enrich Wikipedia articles and make the articles more clear and relevant. In addition, we add a variant of GoPPE (\textit{GoPPE-NoAtt.}) where we replace node attributes with pure Gaussian noise, demonstrating the effectiveness of the positional encodings.


\paragraph{\textsc{GoPPE} is robust against noisy node attributes and consistently outperforms the baselines}: 
Table \ref{tab:exp-2-noisy} presents the average accuracy over snapshots where the node features are gradually refined as defined in Equ.\ref{equ:noisy}.
As expected, the models without positional encodings (MLP, PPRGo, and InsGNN) perform poorly as they heavily depend on the node attributes.  
Meanwhile, \textsc{GoPPE} and its variant (GoPPE-NoAtt) demonstrate the robustness by having positional encodings and outperforms others. 

\begin{table}[ht]
    \centering
    \caption{GoPPE is robust to noisy node attributes. 
    We exclude $\textsc{DynPPE}$ as it is an attribute-free algorithm and has the same results as reported in Table \ref{tab:clf-exp-1}, instead we use GoPPE (NoAtt.) as the baseline where we replace the node attributes with Gaussian noise.
    }
    \vspace{-3mm}
    \label{tab:exp-2-noisy}

    \xingzhiswallow{
    \begin{tabular}{lrrrrr}
    \toprule
    methods &   cora &  citeseer &  pubmed &  arxiv &  flickr \\
    \midrule
    MLP      & 0.2905 &    0.2624 &  0.4816 &      0.2503 &  0.3941 \\
    PPRGo    & 0.3967 &    0.3471 &  0.5837 &      0.3599 &  0.4102 \\
    InsGNN   & 0.4000 &    0.3562 &  0.6378 &      0.4324 &  0.4062 \\ \hline
    GoPPE    & \textbf{0.7797} &    \textbf{0.6307} &  \textbf{0.7745} &      \textbf{0.4636} &  \textbf{0.4338} \\
        \bottomrule
    \end{tabular}
    }
\xingzhiswallow{
    \begin{tabular}{lrrrrrrrr}
        \toprule
             &   Cora &  Citeseer &  Pubmed &  Arxiv &  Flickr &  BlogCatalog &   DBLP &  Physics \\
        \midrule
        MLP      & 0.2905 &    0.2624 &  0.4816 &      0.2503 &  0.3941 &       0.2046 & 0.6415 &   0.5921 \\
        GCN      & 0.5866 &    0.5582 &  0.7589 &         \textit{OOM} &  0.4332 &       0.3036 & 0.6703 &   0.5970 \\
        PPRGo    & 0.3967 &    0.3471 &  0.5837 &      0.3599 &  0.4102 &       0.1927 & 0.6796 &   0.6092 \\
        InsGNN   & 0.4000 &    0.3562 &  0.6378 &      \textbf{0.4324} &  0.4062 &       0.1782 & 0.6915 &   0.5901 \\ \hline
        GoPPE (No.Att)  & \textbf{0.7853} &    \textbf{0.6232} &  \textbf{0.7678} &      0.2620 &  \textbf{0.4362} &       \textbf{0.5040} & \textbf{0.7367} &   \textbf{0.8828} \\
        GoPPE    & \textbf{0.7797} &    \textbf{0.6307} &  \textbf{0.7745} &      \textbf{0.4636} &  \textbf{0.4338} &       \textbf{0.4914} & \textbf{0.7257} &   \textbf{0.8842} \\
        \bottomrule
        \end{tabular}

}
\small
\begin{tabular}{lrrrr|R{10mm}r}
\toprule
 &    MLP &    GCN &  PPRGo &  InsGNN  &  GoPPE (NoAtt.) &  GoPPE \\
\midrule
Cora        & 0.2905 & 0.5866 & 0.3967 &  0.4000 &               \textbf{0.7853} & \textbf{0.7797} \\
Citeseer    & 0.2624 & 0.5582 & 0.3471 &  0.3562 &               \textbf{0.6232} & \textbf{0.6307} \\
Pubmed      & 0.4816 & 0.7589 & 0.5837 &  0.6378 &               \textbf{0.7678} & \textbf{0.7745} \\
Arxiv  & 0.2503 &    \textit{OOM} & 0.3599 &  \textbf{0.4324} &               0.2620 & \textbf{0.4636} \\
Flickr      & 0.3941 & 0.4332 & 0.4102 &  0.4062 &               \textbf{0.4362} & \textbf{0.4338} \\
BlogCatalog & 0.2046 & 0.3036 & 0.1927 &  0.1782 &               \textbf{0.5040} & \textbf{0.4914} \\
DBLP        & 0.6415 & 0.6703 & 0.6796 &  0.6915 &               \textbf{0.7367} & \textbf{0.7257} \\
Physics     & 0.5921 & 0.5970 & 0.6092 &  0.5901 &               \textbf{0.8828} & \textbf{0.8842} \\
\bottomrule
\end{tabular}

\end{table}


\begin{table}[ht]
    \centering
    \caption{We present the average accuracy across snapshots in the intact case. $\textsc{GoPPE}$ is comparable to or better than the baselines while it's the fastest as discussed in Section \ref{sec:ppr-speed}.  
    }
    \vspace{-3mm}
    \label{tab:clf-exp-1}

    \xingzhiswallow{
        \begin{tabular}{l|rrrrr}
        \toprule
        models  &   cora &  citeseer &  pubmed &  arxiv &  flickr \\
        \midrule
        Random   & 0.1977 &    0.1667 &  0.3731 &      0.1343 &  0.3530 \\
        MLP      & 0.4886 &    0.3987 &  0.7098 &      0.3102 &  0.4204 \\
        PPRGo    & 0.7101 &    0.5458 &   \textbf{0.8139} &      0.4244 &  0.4128 \\
        InsGNN   & 0.6925 &    0.5588 &  \textbf{0.8362} &      \textbf{0.4772} &  0.4138 \\
        DynPPE   & \textbf{0.7984} &     \textbf{0.6314} &  0.7821 &       0.4648 &  \textbf{0.4220} \\ \hline
        GoPPE    &  \textbf{0.7935} &     \textbf{0.6310} &   0.7861 &       \textbf{0.4843} &   \textbf{0.4427} \\
            \bottomrule
        \end{tabular}
    }

\xingzhiswallow{
    \begin{tabular}{lrrrrrrrr}
\toprule
         &   cora &  citeseer &  pubmed &  ogbn-arxiv &  flickr &  BlogCatalog &   DBLP &  Physics \\
    \midrule
    MLP      & 0.4886 &    0.3987 &  0.7098 &      0.3102 &  0.4204 &       0.2607 & 0.6401 &   0.6452 \\
    GCN      & 0.7134 &    0.6222 &  0.7725 &         \textit{OOM} &  \textbf{0.4440} &       0.3782 & 0.6929 &   0.6637 \\
    PPRGo    & 0.7101 &    0.5458 &  \textbf{0.8139} &      0.4244 &  0.4128 &       0.2139 & 0.6925 &   0.6647 \\
    InsGNN   & 0.6925 &    0.5588 &  \textbf{0.8362} &      \textbf{0.4772} &  0.4138 &       0.1782 & 0.7025 &   0.6703 \\ 
    DynPPE   & \textbf{0.7984} &    \textbf{0.6314} &  0.7821 &      0.4648 &  0.4220 &       \textbf{0.5112} & \textbf{0.7264} &   \textbf{0.8762} \\  
    \hline
    GoPPE    & \textbf{0.7935} &    \textbf{0.6310} &  0.7861 &      \textbf{0.4843} &  \textbf{0.4427} &       \textbf{0.4941} & \textbf{0.7317} &   \textbf{0.8792} \\
    \bottomrule
    \end{tabular}

}

\small
\begin{tabular}{lrrrrr|r}
\toprule
 &    MLP &    GCN &  PPRGo &  InsGNN &  DynPPE &   GoPPE \\
\midrule
Cora        & 0.4886 & 0.7134 & 0.7101 &  0.6925 &  \textbf{0.7984} &      \textbf{0.7935} \\
Citeseer    & 0.3987 & 0.6222 & 0.5458 &  0.5588 &  \textbf{0.6314} &       \textbf{0.6310} \\
Pubmed      & 0.7098 & 0.7725 & \textbf{0.8139} &  \textbf{0.8362} &  0.7821 &       0.7861 \\
Arxiv  & 0.3102 &    \textit{OOM} & 0.4244 &  0.4772 &  \textbf{0.4648} &    \textbf{0.4843} \\
Flickr      & 0.4204 & \textbf{0.4440} & 0.4128 &  0.4138 &  0.4220 &         \textbf{0.4427} \\
BlogCatalog & 0.2607 & 0.3782 & 0.2139 &  0.1782 &  \textbf{0.5112} &        \textbf{0.4941} \\
DBLP        & 0.6401 & 0.6929 & 0.6925 &  0.7025 &  \textbf{0.7264} &        \textbf{0.7317} \\
Physics     & 0.6452 & 0.6637 & 0.6647 &  0.6703 &  \textbf{0.8762} &        \textbf{0.8792} \\
\bottomrule
\end{tabular}

\end{table}

\paragraph{GoPPE achieves comparable performance with intact features }:  Table \ref{tab:clf-exp-1} present the averaged prediction accuracy across all snapshots.  
Specifically, we found the positional encoding (\textsc{DynPPE}) alone can yield strong performance, further demonstrating the importance of locality information. 
 Meanwhile, we will analyze the performance difference in \textit{pubmed} and \textit{flickr} later in Section \ref{sec:discuss}.

\subsection{Discussions and Future Works} 
\label{sec:discuss}
\paragraph{Robust positional encodings achieves better accuracy over graphs of \textit{weak} node attributes }: Table \ref{tab:clf-exp-1} shows that \textsc{MLPs} provides poor performance on \textit{cora} and \textit{citeseer}, implying that these graphs have weak node attributes for prediction. It makes \textsc{PPRGo} and  \textsc{InsGNN} harder to extract signals from attributes. Meanwhile, the positional encodings successfully capture the node context and achieve better performance.   
On the other hand, on the attribute-rich graphs, even \textsc{MLP} performs relatively well, for example, on \textit{pubmed} graph. It is because \textit{pubmed} has dense and informative node attributes, which makes the PPR-based feature aggregation (\textsc{PPRGo},\textsc{InsGNN}) produce a succinct and predictive signal. 



\paragraph{How to adapt PPR propagation over dynamic heterotrophic graph remains an open problem}: 
Besides the attribute-rich property in \textit{pubmed},  it is also known for its homophily property \cite{mcpherson2001birds} which makes the nodes of the same labels well-connected.  
As we discussed in the Definition \ref{def:att-property}, one of the assumptions is graph homophily, which makes PPR-based aggregation predictive with local \textit{peer} nodes.  However, \textit{flickr} is a heterotrophic graph \cite{kim2022find}. Table \ref{tab:clf-exp-1} shows that the  accuracy of all models is worse or just slightly better than \textsc{MLP}. 
Inspired by \citet{chien2020adaptive}, we are interested in adapting PPR for dynamic heterotrophic graphs.

\paragraph{Faster PPR-GNN with efficient PPR solver}:
Given the impressive efficiency of ISTA-based PPR solver, it motivates us to explore more toward faster optimization methods for dynamic PPR settings. For example, FISTA or Blockwise Coordinate Descent methods with faster rule and active set \cite{nutini2017let}.  Such research directions have the potential for even faster GNN designs for dynamic graphs and tackle larger-scale problems.

\section{Conclusion }
\label{sec:conclusion}
In this paper, we proposed an efficient framework for node representation learning over dynamic graphs. Within the framework, we propose the desired properties in node attention design for global feature aggregation, and we justify the use of PPR under its $\ell_1$-regularized formulation. Besides, we proposed to use ISTA as the PPR solver and use warm-start as an elegant solution to maintain PPR over dynamic graphs. Finally, we introduce the PPR-based node positional encodings, maximizing the usage of PPR for robust representation against noisy node attributes.
The experiments show that our instantiated model \textsc{GoPPE} is upto 6 times faster than the current state-of-the-art models while achieving comparable or better prediction accuracy over graphs with intact node attributes. In the noisy environment, \textsc{GoPPE} significantly outperforms the others, demonstrating the effectiveness of the proposed positional encodings, benefiting the downstream tasks in knowledge discovery \cite{wang2022knowledge, guo2022hierarchies, wang2023knowledge, wang2023knowledge2, sultan2022low, guo2019inferring}, anomaly detection \cite{zhang2023subanom} and recommendation system \cite{lin2023comet} in the dynamic environment.


\xingzhiswallow{
In this paper, we propose a unified framework \textsc{DynAnom} for subset node anomaly tracking over large dynamic graphs. This framework can be easily applied to different graph anomaly detection tasks from local to global with a flexible score function customized for various applications. 
Experiments show that our proposed framework outperforms current state-of-the-art methods by a large margin, and has a significant speed advantage (2.3 times faster) over large graphs.  
We also present a real-world PERSON graph with an interesting case study about personal life changes, providing a rich resource for both knowledge discovery \cite{guo2022hierarchies} and algorithm benchmarking.
For future work, it remains interesting to explore a different type of score functions, automatically identify the interesting subset of nodes as better strategies for tracking global-level anomaly, and further investigate temporal-aware PageRank as better node representations.
}


\appendix
\section{Appendix}
\label{app:algos-proofs}

\subsection{Forward Push Algorithm}

\begin{algorithm}[ht]
\caption{$\textsc{Forward Push}$}
\begin{algorithmic}[1]
\State \textbf{Input: }$\bm \pi_s, \bm \pi_s, \mathcal{G}, \epsilon, \alpha$
\While{ exists $i$ such that $| r_s(i)| > \frac{\epsilon}{m} d(i)$}
\State $\textsc{Push}(i)$
\EndWhile
\State \Return $(\bm \pi_s, \bm r_s)$
\Procedure{Push}{$i$}
\State $\pi_s(i) \pluseq \alpha r_s(i)$
\For{$j \in \operatorname{Nei}_{out}(i)$}
\State $r_s(j) \pluseq \frac{(1-\alpha)  r_s(i)}{ \bm d_{out}(i) } $
\EndFor
\State $r_s(i) = 0 $
\EndProcedure
\end{algorithmic}
\label{algo:local-push}
\end{algorithm}


\begin{algorithm}[ht]
\caption{$\textsc{GoPPE}(\mathcal{G}_0, \Delta E_{1, \ldots, T}, \bm X_{1, \ldots, T}, \mathcal{S}, \epsilon, \alpha)$ }
\begin{algorithmic}[1]
\State \textbf{Input: }  
Initial graph $\mathcal{G}_0 = (\mathcal{V}, \mathcal{E}_0, \bm X_0)$, 
Edge events $ \Delta E_{1, \ldots, T}$, 
Node attributes $ \bm X_{1, \ldots, T}$ , 
The train/test node sets, denoted as $\mathcal{\bar{S}} \text{ and }\mathcal{\hat{S}}$, respectively. And $\mathcal{S} = \mathcal{\bar{S}} \cup \mathcal{\hat{S}}$. 
PPR teleport factor $\alpha $,
PPR Vector $\ell_1$-error control parameter $\epsilon$.

\State $\bm \pi_{u, init} = \bm 0, \forall u \in \mathcal{S}$
\State \textcolor{blue}{// Calculate Initial PPR for each node $u$ in $\mathcal{S}$ using Algorithm \ref{algo:ppr-ista}.}
\State $\bm x_{u, t=0}, \nabla_{u, t=0} = \textsc{PPRIsta}(\mathcal{G}_0, \bm \pi_{u, init}),  \forall u \in \mathcal{S} $ 
\State $\bm \pi_{u, t=0} =  \sqrt{\bm d(u)} \times \bm x_{u, t=0}  ,  \forall u \in \mathcal{S} $
\State \textcolor{blue}{// Get Contextualized Node Features and Positional Encodings}
\State $\bm h_{u,t=0} = \textsc{FeatureFormer}(\bm \pi_{u, t=0},\bm X_0),  \forall u \in \mathcal{S} $ 
\State \textcolor{blue}{// Train classifier with the training samples $\bar{u} \in \mathcal{\bar{S}} \subset \mathcal{S} $}
\State $f_{\bm \Theta,t=0}(\cdot) = \textsc{Train}(\bm h_{\bar{u}, t=0}, \bm y(\bar{u})),  \forall \bar{u} \in \mathcal{\bar{S}}$ 
\State \textcolor{blue}{// Infer for the testing samples $\bm h_{\hat{u},t=0}, \forall \hat{u} \in \mathcal{\hat{S}}$}
\State $ \bar{y}_{\hat{u}, t=0} = f_{\bm \Theta,t=0}(\bm h_{\hat{u}}), \forall \hat{u} \in \mathcal{\hat{S}} $ 
\State \textcolor{blue}{// Update graphs, maintain PPR, then train/evaluate repeatedly.}
\For{$t \in [1,T] $}
    \State $ \bar{\bm x}_{u, t-1} =\textsc{PPRAdjust}(\bm x_{u, t-1}, \Delta E_{t})$
    \State $\mathcal{G}_{t} = \textsc{GraphUpdate}(\mathcal{G}_{t-1}, \Delta E_t, \bm X_t)$
    \State $\bm x_{u, t}, \nabla_{u, t} = \textsc{PPRIsta}(\mathcal{G}_1, \bar{\bm x}_{u, t-1}),  \forall u \in \mathcal{S} $ 
    \State $\bm \pi_{u, t} =  \sqrt{\bm d(u)} \times \bm x_{u, t}  ,  \forall u \in \mathcal{S} $
    \State $\bm h_{u,t} = \textsc{FeatureFormer}(\bm \pi_{u, t},\bm X_t),  \forall u \in \mathcal{S} $
    \State $f_{\bm \Theta,t}(\cdot) = \textsc{Train}(\bm h_{\bar{u}, t}, \bm y(\bar{u})),  \forall \bar{u} \in \mathcal{\bar{S}}$ 
    \State $ \bar{y}_{\hat{u},t} = f_{\bm \Theta,t}(\bm h_{\hat{u}}), \forall \hat{u} \in \mathcal{\hat{S}} $ 
\EndFor

\State \textbf{return} $\bar{y}_{\hat{u},t \in \{ 0, 1, \ldots, T \}}, \forall \hat{u} \in \mathcal{\hat{S}}$

\noindent\hrulefill
\Procedure{FeatureFormer}
{$\bm \pi_u,\bm X$}
\State \textcolor{blue}{// PPR-guided Node Contextualization}
\State $\bm h^c_u =  \bm X \bm \pi_u $
\State \textcolor{blue}{// PPR-based Node Positional Encodings}
\State $\bm p_i = \textsc{HashReduceDim}(\bm \pi_u, \bar{d}=512)$
\State $\bm h^p_u = \bm W_{pe} \bm p_i $
\State \textbf{return}  Apply \textit{Concatenative PE }in Equ.\ref{equ:goppe} 
\EndProcedure

\noindent\hrulefill
\Procedure{PPRAdjust}{$ \bm x, \Delta E$}
\For{$(u ,v)\in \Delta E$}
    \State $\bar{\bm x}_u:=$ update $\bm x_u$ Using Theorem \ref{theorem:update}.
\EndFor

\State \textbf{return} $ \bar{\bm x}$ 
\EndProcedure

\noindent\hrulefill
\Procedure{HashReduceDim}{$x,d$ } \label{proc:HashReduce}
\State // Hash function $h_{d}(i): \mathbb{N}\to [d]$
\State // Hash function $h_{\operatorname{sgn}}(i): \mathbb{N}\to \{ \pm 1\}$
\State $\bar x = \bm 0 \in \mathbb{R}^{d}$
\For{$i \in \textsc{Supp}|x|$}
    \State $j = h_{dim}(i)$
    \State $\bar x(j) \pluseq h_{\operatorname{sgn}}(i) \log \left(x(i) \right)$
\EndFor
\State \textbf{return} $ \frac{\bar x}{\| \bar x \|_1}$
\EndProcedure

\end{algorithmic}
\label{algo:goppe}
\end{algorithm}


\xingzhiswallow{
\subsection{PPR Invariant Property}
\paragraph{PPR Invaraint over dynamic graphs} 
\citet{zhang2016approximate} and \citet{xingzhi2021subset} proposed the incremental update rules for PPR over dynamic graphs.
 
\begin{lemma} [PPR Invariant Property \cite{zhang2016approximate}] Suppose $\bm \pi_s$ is the PPV of node $s$ on graph $\mathcal{G}_t$. Let $\bm p_s$ and $\bm \pi_s$ be returned by the weighted version of \textsc{DynamicForwardPush} presented in Algo. \ref{algo:forward-local-push}. Then, we have the following invariant property.
\begin{align}
\pi_s(u) &= p_s(u) + \sum_{x \in \mathcal{V}}r_s(x) \pi_s(x),  \text{ for all } u \in \mathcal{V},  \nonumber \allowdisplaybreaks\\
p_s(u) + \alpha r_s(u) &= (1-\alpha) \sum_{x \in N^{in}(u)} \frac{w_{(u,x)}p_s(x)}{d(x)} + \alpha 1_{u=s},\nonumber \allowdisplaybreaks
\end{align}
where $N^{in}(\cdot)$ is the in-neighbors, and $1_{u=s} = 1$ if $u = s$, 0 otherwise.
\label{lemma:ppr-invar}
\end{lemma}
}

\xingzhiswallow{

\subsection{Formulate Quadratic Objective from  Forward Push }
\label{sec:proof-push-opt}

\xingzhiswallow{

\begin{lemma}[$\ell_1$-regularized Quadratic Optimization Problem of PPR ] 
\todo{add-ref}Applying the Forward Push algorithm is equivalent to applying the Coordinate Descent algorithm to optimize the following objective function in quadratic form: 

\begin{align}
    \bm x^{*} &= \argmin_{\bm x } \left( \frac{1}{2} \bm x^{\top} \bm W \bm x - \bm (\alpha \bm D^{-1/2} \bm e_s)^{\top} \bm x  \right), where \label{eq:ppr-quad-form-app}\\
    \bm W &= \bm D^{-1/2}(\bm D - (1-\alpha) \bm A ) \bm D^{-1/2}\\
    \bm \pi &= \bm D^{1/2} \bm x^{*}
\end{align}


\end{lemma}
}

\paragraph{From Push to the objective function }: The proofs follows \cite{fountoulakis2019variational}.
One can rewrite Forward Push algorithm into $\textsc{CD}$:

Given Equ.\ref{equ:ppr}, we define Residual Vector $\bm r \in \mathbb{R}^{n}$ as followings:
\begin{align}
    \bm r := (\bm I_n - (1-\alpha) \bm A \bm D^{-1} ) \bm \pi - \alpha \bm e_s  \label{eq:push-update}
\end{align}

By multiplying $\bm D^{-1/2}$ on both sides of Equ.\ref{eq:push-update} and  $\bm \pi := \bm D^{1/2} \bm x$:
\begin{align}
    \bm D^{-1/2} \bm r &= \bm W \bm x + \bm b,   
    \label{eq:grad-q}\\
    where \ \bm W &= \bm D^{-1/2}(\bm D - (1-\alpha) \bm A) \bm D^{-1/2}, \nonumber \\
    \bm b &= - \alpha \bm D^{-1/2} \bm e_s \nonumber
\end{align}

Equ.\ref{eq:grad-q} has a linear form w.r.t. $\bm x$ which can be interpreted as the gradient $\nabla f(\bm x) := \bm D^{-1/2} \bm r = \bm W \bm x + \bm b $ of a quadratic function $f(\cdot)$ such that:
\begin{align}
    f(\bm x) &:= \frac{1}{2} \bm x^{\top} \bm W \bm x + \bm b^{\top} \bm x  , \label{eq:ppr-quad-form-app2}
\end{align}

The most intriguing finding is that by applying coordinate descent to Equ.\ref{eq:ppr-quad-form} and minimizing the objective function, the solution $\bm x^{*} = \operatorname{argmin}_{\bm x}f(\bm x)$ is the approximate PPR given the fact that $\bm \pi = \bm D^{1/2} \bm x^{*}$. Such formulation bridges two seemingly disjoint fields and provides more opportunities to solve PPR through the lens of optimization. 

\paragraph{Rewrite Coordinate Descent for PPR}

Note that $\bm r = \bm D^{1/2}\nabla f(\bm x) $, $\bm \pi = \bm D^{1/2} \bm x,$ 

\begin{algorithm}[ht]
\caption{$\textsc{CD Solver for PPR}$ }
\begin{algorithmic}[1]
\State \textbf{Input: }$ k = 0, \bm x^{(0)} = \bm 0, \mathcal{G}, \epsilon, \alpha$
\State $\nabla f(\bm x^{(0)}) = - \alpha \bm D^{-1/2} \bm e_s$

\While{ exists $i$ such that $\|\nabla f(\bm x)(i)\|_1 > \epsilon' d(i)^{1/2}$}

\State $\textsc{CD}(i)$
\EndWhile
\State \Return $\left(\bm \pi = \bm D^{1/2} \bm x^{(k)}, \bm r =  \bm D^{1/2}\nabla f(\bm x^{(k)}) \right)$
\Procedure{CD}{$i$}
\State  \textcolor{blue}{/*apply coordinate descent*/} 
\State $\bm x^{(k+1)}(i) =  \bm x^{(k)}(i) -  \nabla_i f(\bm x^{(k)}) $
\State  \textcolor{blue}{/*update gradient*/} 
\State $\nabla f(\bm x^{(k+1)})(i) = 0$
\For{$j \in \operatorname{Nei}_{out}(i)$}
\State $  \bm \nabla_j f(\bm x^{(k+1)})  = \nabla_j f(\bm x^{(k)})  + \frac{(1-\alpha) \bm A_{(i,j)}  }{\bm d(i)^{1/2}\bm d(j)^{1/2}} \nabla_i f(\bm x^{(k)}) $
\EndFor
\State $k = k+1$
\EndProcedure
\end{algorithmic}
\label{algo:forward-local-push}
\end{algorithm}

By looking closely to the termination condition: 
$\|\nabla f(\bm x)(i)\|_1 \leq \epsilon' d(i)^{1/2} 
\Rightarrow 
\nabla f(\bm x)(i) \geq - \epsilon' d(i)^{1/2}$ 
given $\nabla f_i(\bm (x) \leq 0, \forall i$, 
\xingzhiswallow{

\begin{lemma}[PPR as a $\ell_1$-regularized Quadratic Optimization]
the termination condition is equivalent to the first-order optimality condition of the following objective function with $\ell_1$-regularization:

\begin{align}
    \min_{\bm x} \psi(\bm x) := f(\bm x) + \epsilon'\| \bm D^{1/2} \bm x \|_1
\end{align}

    
\end{lemma}

}

Note that the optimality conditions imply the termination condition, and $\nabla_i f(\bm x^{*}) \in [ -\epsilon' \bm d(i)^{1/2},0 ] \text{ if }  \bm x^{*}(i) = 0 $ implies its sparsity.

$\operatorname{prox}(\bm x) = \operatorname{argmin}_{\bar{\bm x}} \frac{1}{2}\|  \bm x - \bar{\bm x} \|_2^2 + \epsilon' \|\bm D^{1/2} \bm x \|_1 $

}

\xingzhiswallow{
\subsection{PPV maintenance on dynamic graphs}
\label{proof:ppv-maintain}
From \cite{zhang2016approximate}: Given a new edge $e_{u,v}$, the adjustments are :
\begin{align}
    \bm \pi'(u) &= \bm \pi(u) * \frac{\bm d(u)+1}{\bm d(u)}\\
    \bm r'(u) &= \bm r(u) - \frac{\bm \pi(u)}{ \alpha \bm d(u)}\\
    \bm r'(v) &= \bm r(v) + \frac{1-\alpha}{\alpha} \frac{\bm \pi(u)}{ \bm d(u)}
\end{align}

Note that, in PPR-equivalent quadratic formulation, we define:
\begin{align}
     \bm r = \bm D^{1/2} \nabla f(\bm x) \text{ and } \bm \pi = \bm D^{-1/2} \bm x 
\end{align}
In addition, the PPR invariant property still holds for the results from the solution.
When a new edge event arrives, the update rules adjust $\bm x$ and $\nabla f(x)$ for further updating. We substitute them back to the update rules and have:

\begin{align}
    \bm d^{-1/2}(u) * \bm x'(u) &= \bm d^{-1/2}(u) * \bm x(u) * \frac{\bm d(u)+1}{\bm d(u)} \nonumber \\
    \Rightarrow \bm x'(u) &= \bm x(u) * \frac{\bm d(u)+1}{\bm d(u)}\nonumber \\
    \bm d^{1/2}(u) * \nabla f'(\bm x')(u) &= \bm d^{1/2}(u)* \nabla f(\bm x)(u) - \frac{\bm d^{-1/2}(u) * \bm x(u)}{ \alpha \bm d(u)}\nonumber \\
    \Rightarrow \nabla f'(\bm x')(u) &= \nabla f(\bm x)(u) - \frac{\bm x(u)}{ \alpha \bm d^{1/2}(u)}\nonumber \\
    \bm d^{1/2}(v) * \nabla f'(\bm x')(v) &= \bm d^{1/2}(v)*\nabla f(\bm x)(v)  + \frac{1-\alpha}{\alpha} \frac{\bm d^{1/2}(u) * \bm x(u)}{ \bm d(u)}\nonumber \\
     \Rightarrow \nabla f'(\bm x')(v) &= \nabla f(\bm x)(v)  + \frac{(1-\alpha) }{\alpha} \frac{ \bm d^{-1/2}(v)*\bm x(u)}{ \bm d^{1/2}(u)}\nonumber 
\end{align}

}

\xingzhiswallow{
    \subsection{Dimension Reduction using Random Projection }
    The sparse Gaussian random projection can be constructed:
    \begin{align}
        \bm p_i = \bm R \bm \pi_i, \text{ and }
        \bm R \in \mathbb{R}^{d_i \times |\mathcal{V}|}  := 
        \begin{dcases}
            \small
            +\sqrt{3} \text{ with probability } \frac{1}{6}\\
            \sqrt{0} \text{ with probability } \frac{2}{3}\\
            -\sqrt{3} \text{ with probability } \frac{1}{6}\\
        \end{dcases}
        \label{equ:rand-proj}
    \end{align}

    }
\subsection{Experiment Details and Reproducibility}
\label{app:train-detail}

\paragraph{Infrastructure:} 
All the experiments are done on a machine equipped with 4 processors of Intel(R) Xeon(R) CPU E5-2630 v4 @ 2.20GHz, where each processor has 10 cores, resulting in 40 cores in total and  125GB main memory. Besides, we have 4 Nvidia-Titan-XP GPUs where each has 12GB GPU memory.

\paragraph{PPR Precision Control:} 
We maintain a similar precision level (with $\epsilon \approx 1e-8$ for graphs, except for Arxiv graph with $\epsilon \approx 1e-6$ ) across all graphs. The error was calculated against the power iteration results with a smaller $\epsilon$ setting for higher precision.

\xingzhiswallow{

    \begin{table}[ht]
        \centering
        \caption{We maintain a similar precision level ($\epsilon \approx 1e-6$ or $1e-8$ ) across all graphs. The error was calculated against the power iteration results with a smaller $\epsilon$ setting for higher precision.}
        \label{tab:ppr-acc}
        \footnotesize
    \begin{tabular}{llrrrrr}
    \toprule
    ppr-algo &  increment &     cora &  citeseer &   pubmed &  ogbn-arxiv &   flickr \\
    \midrule
        push &      False & 3.99e-08 &  2.25e-08 & 4.57e-08 &    4.76e-06 & 4.32e-08 \\
        ista &      False & 5.22e-08 &  2.78e-08 & 6.03e-08 &    6.08e-06 & 2.84e-08 \\
        push &       True & 4.06e-08 &  2.30e-08 & 4.65e-08 &    4.85e-06 & 4.34e-08 \\
        ista &       True & 5.14e-08 &  2.73e-08 & 5.94e-08 &    6.00e-06 & 2.80e-08 \\
    \bottomrule
    \end{tabular}
    \end{table}
    
}

\xingzhiswallow{
    \begin{table}[ht]
        \centering
        \footnotesize
        \begin{tabular}{llrrrr}
        \toprule
        ppr-algo &  increment &     cora &  citeseer &   pubmed &   flickr \\
        \midrule
            push &      False & 3.99e-08 &  2.25e-08 & 4.57e-08 & 4.32e-08 \\
            ista &      False & 5.22e-08 &  2.78e-08 & 6.03e-08 & 2.84e-08 \\
            push &       True & 4.06e-08 &  2.30e-08 & 4.65e-08 & 4.34e-08 \\
            ista &       True & 5.14e-08 &  2.73e-08 & 5.94e-08 & 2.80e-08 \\
        \bottomrule
        \end{tabular}
        \caption{The PPR precision}
        \label{tab:my_label}
    \end{table}
    
    }

\paragraph{PPR Implementation} We implemented all PPR routines (Power iteration, Forward Push (w/o dynamic adjustment), ISTA) in Python and ensured that all are Numba-accelerated ($\operatorname{@njit}$). Numba acceleration compiles python routines to LLVM native code, which has a comparable speed to C/C++. 
We particularly chose Python+Numba as the main programming tool because it's easier to maintain and is gradually being accepted by the scientific computing community.

\paragraph{Neural Model Configurations }
We implemented the neural networks using PyTorch. 
We use $\textsc{Linear}\rightarrow\textsc{Relu} \rightarrow\textsc{Dropout}$ as the basic neural compoent for classification. 
Besides. we set dropout ratio $p=0.15$ during training.
During training, we use \textsc{Adam} optimizer and force all models to train for 50/100 epochs before early exit.
More details can be found in the training scripts in our codebase included in this submission.
To have a fair comparison, we use the same model configurations except for the proposed feature aggregation function. Note that we denote the dimension of the node's raw feature as $|Feat|$ and the total node label as $|L|$. We use the presented hidden state sizes (128, 32, 16) in MLP-\{0,1,2\}  for small graphs, while use larger size (128, 512, 256) for larger graph. Likewise, all methods share the same architecture configs.

\xingzhiswallow{

\begin{table}[!h]
\centering
\footnotesize
\caption{To have a fair comparison, we use the same model configurations except for the proposed feature aggregation function. Note that we denote the dimension of the node's raw feature as $|Feat|$ and the total node label as $|L|$. We use the presented hidden state sizes (128, 32, 16) in MLP-\{0,1,2\}  for small graphs, while use larger size (128, 512, 256) for larger graph. Likewise, all methods share the same architecture configs. }
\label{tab:model-param}
\begin{tabular}{|p{0.06\textwidth}|p{0.05\textwidth}|p{0.05\textwidth}|p{0.05\textwidth}|p{0.06\textwidth}|p{0.05\textwidth}|}
\hline 
  & MLP-Feat & MLP-PPE & MLP-0 & MLP-1 & MLP-2 \\ \hline 
 Gaussian & |Feat| * 128 & - & 128*32 & 32*16 & 16*|L| \\
\hline 
 PPE & - & |PPE| * 128 & 128*32 & 32*16 & 16*|L| \\
\hline 
 MLP & |Feat| * 128 & - & 128*32 & 32*16 & 16*|L| \\
\hline 
 PPRGo & |Feat| * 128 & - & 128*32 & 32*16 & 16*|L| \\
\hline 
 InstantGNN & |Feat| * 128 & - & 128*32 & 32*16 & 16*|L| \\
\hline 
 \textbf{GoPPE} & |Feat| * 128 & |PPE| * 128 & 128*32 & 32*16 & 16*|L| \\
 \hline
\end{tabular}
\end{table}
}


\xingzhiswallow{
\subsection{Reproducibility}
We repeated the experiments with three different random seeds for all stochastic parts (e.g., python, numpy and PyTorch) during training and testing. We included the source code and the scripts needed to reproduce the results and will release them to the public upon publication.
}

\subsection{Edge event sampling}
We create dynamic graphs from static graphs. First, we associate each edge event with a distinctive timestamp from 1 to $|\mathcal{E}|$. Then, we use the first edges $\Delta\mathcal{E}_0$ (e.g., with timestamp < 100) to construct $\mathcal{G}_0$ as the base graph. Finally, we inject a fixed number of edges  $\Delta\mathcal{E}_1$ (e.g. with the timestamp from 100 to 120) to simulate graph evolution to create the next graph snapshot $\mathcal{G}_1 = \mathcal{G}_0 + \Delta\mathcal{E}_1$. Iteratively, we create all snapshots. Note that the created dynamic graph may not have temporal characteristics and temporal modeling is not in the scope of this paper.

\xingzhiswallow{
\subsection{Node sampling}
We split data into train/dev/test sets using stratified sampling w.r.t. the node label. To avoid undefined metrics in evaluation, we ensure that each node label has at least one sample appearing in the train, dev, and test sets (a total of 3 different samples).

}

\xingzhiswallow{
\subsection{Extra results}

\begin{figure*}[htbp]
    \includegraphics[width=1.0\textwidth]{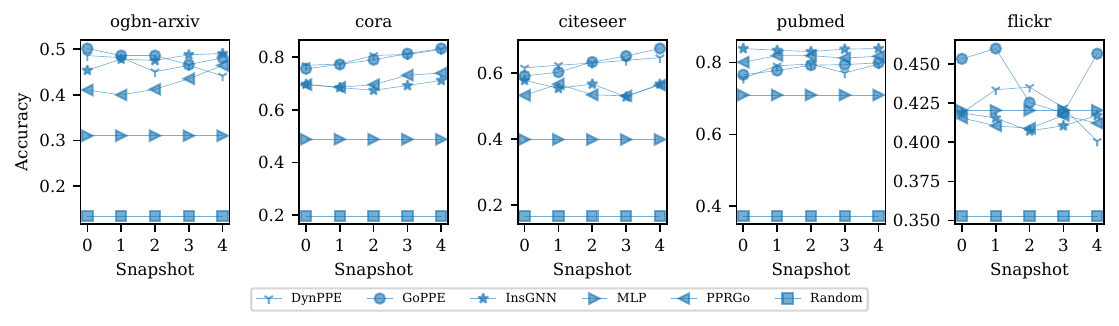}
    \caption{Prediction accuracy on intact features across snapshots. \textsc{GoPPE} has comparable or better accuracy than other baselines.}
    \label{fig:exp-1-clf}
\end{figure*}

\begin{figure*}[htbp]
    \includegraphics[width=1.0\textwidth]{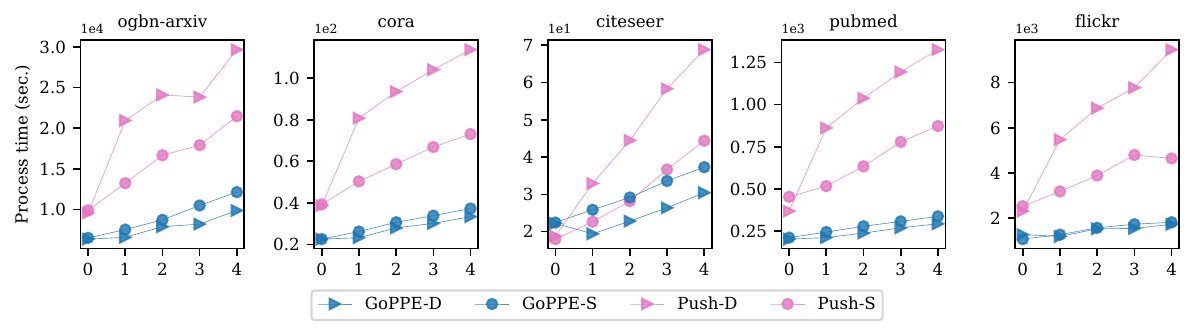}
    \caption{PPR calculation time in major change case.}
    \label{fig:exp-3-ppr-time-major}
\end{figure*}

\begin{figure*}[htbp]
    \includegraphics[width=1.0\textwidth]{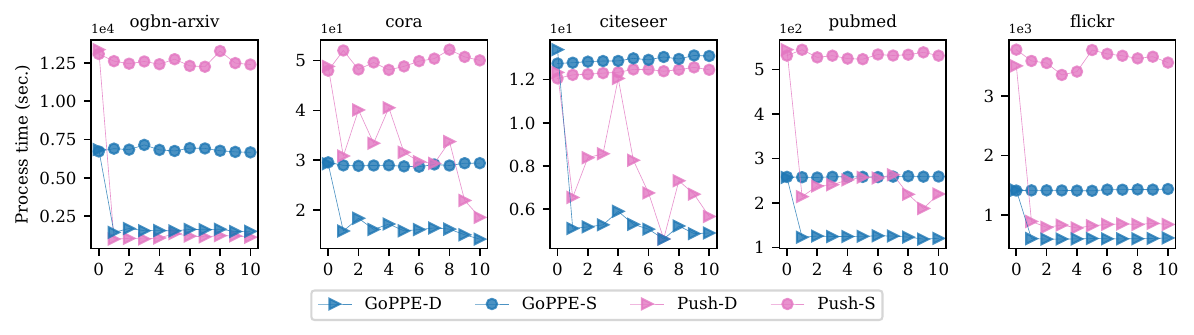}
    \caption{PPR calculation time in minor change case.}
    \label{fig:exp-3-ppr-time-minor}
\end{figure*}

}


\newpage
\bibliographystyle{ACM-Reference-Format}
\balance
\bibliography{references}


\begin{thebibliography}{54}


\ifx \showCODEN    \undefined \def \showCODEN     #1{\unskip}     \fi
\ifx \showDOI      \undefined \def \showDOI       #1{#1}\fi
\ifx \showISBNx    \undefined \def \showISBNx     #1{\unskip}     \fi
\ifx \showISBNxiii \undefined \def \showISBNxiii  #1{\unskip}     \fi
\ifx \showISSN     \undefined \def \showISSN      #1{\unskip}     \fi
\ifx \showLCCN     \undefined \def \showLCCN      #1{\unskip}     \fi
\ifx \shownote     \undefined \def \shownote      #1{#1}          \fi
\ifx \showarticletitle \undefined \def \showarticletitle #1{#1}   \fi
\ifx \showURL      \undefined \def \showURL       {\relax}        \fi
\providecommand\bibfield[2]{#2}
\providecommand\bibinfo[2]{#2}
\providecommand\natexlab[1]{#1}
\providecommand\showeprint[2][]{arXiv:#2}

\bibitem[Andersen et~al\mbox{.}(2006)]%
        {andersen2006local}
\bibfield{author}{\bibinfo{person}{Reid Andersen}, \bibinfo{person}{Fan Chung}, {and} \bibinfo{person}{Kevin Lang}.} \bibinfo{year}{2006}\natexlab{}.
\newblock \showarticletitle{Local graph partitioning using pagerank vectors}. In \bibinfo{booktitle}{\emph{The IEEE Symposium on Foundations of Computer Science (FOCS)}}. IEEE, \bibinfo{pages}{475--486}.
\newblock


\bibitem[Bai et~al\mbox{.}(2024)]%
        {bai2024faster}
\bibfield{author}{\bibinfo{person}{Jiahe Bai}, \bibinfo{person}{Baojian Zhou}, \bibinfo{person}{Deqing Yang}, {and} \bibinfo{person}{Yanghua Xiao}.} \bibinfo{year}{2024}\natexlab{}.
\newblock \showarticletitle{Faster Local Solvers for Graph Diffusion Equations}.
\newblock \bibinfo{journal}{\emph{arXiv preprint arXiv:2410.21634}} (\bibinfo{year}{2024}).
\newblock


\bibitem[Beck and Teboulle(2009)]%
        {beck2009fast}
\bibfield{author}{\bibinfo{person}{Amir Beck} {and} \bibinfo{person}{Marc Teboulle}.} \bibinfo{year}{2009}\natexlab{}.
\newblock \showarticletitle{A fast iterative shrinkage-thresholding algorithm for linear inverse problems}.
\newblock \bibinfo{journal}{\emph{SIAM journal on imaging sciences}} \bibinfo{volume}{2}, \bibinfo{number}{1} (\bibinfo{year}{2009}), \bibinfo{pages}{183--202}.
\newblock


\bibitem[Bojchevski et~al\mbox{.}(2020)]%
        {bojchevski2020scaling}
\bibfield{author}{\bibinfo{person}{Aleksandar Bojchevski}, \bibinfo{person}{Johannes Klicpera}, \bibinfo{person}{Bryan Perozzi}, \bibinfo{person}{Amol Kapoor}, \bibinfo{person}{Martin Blais}, \bibinfo{person}{Benedek R{\'o}zemberczki}, \bibinfo{person}{Michal Lukasik}, {and} \bibinfo{person}{Stephan G{\"u}nnemann}.} \bibinfo{year}{2020}\natexlab{}.
\newblock \showarticletitle{Scaling graph neural networks with approximate pagerank}. In \bibinfo{booktitle}{\emph{Proceedings of the 26th ACM SIGKDD International Conference on Knowledge Discovery \& Data Mining}}. \bibinfo{pages}{2464--2473}.
\newblock


\bibitem[Brand(2002)]%
        {brand2002incremental}
\bibfield{author}{\bibinfo{person}{Matthew Brand}.} \bibinfo{year}{2002}\natexlab{}.
\newblock \showarticletitle{Incremental singular value decomposition of uncertain data with missing values}. In \bibinfo{booktitle}{\emph{Computer Vision—ECCV 2002: 7th European Conference on Computer Vision Copenhagen, Denmark, May 28--31, 2002 Proceedings, Part I 7}}. Springer, \bibinfo{pages}{707--720}.
\newblock


\bibitem[Chen et~al\mbox{.}(2023)]%
        {chen2023accelerating}
\bibfield{author}{\bibinfo{person}{Zhen Chen}, \bibinfo{person}{Xingzhi Guo}, \bibinfo{person}{Baojian Zhou}, \bibinfo{person}{Deqing Yang}, {and} \bibinfo{person}{Steven Skiena}.} \bibinfo{year}{2023}\natexlab{}.
\newblock \showarticletitle{Accelerating personalized PageRank vector computation}. In \bibinfo{booktitle}{\emph{Proceedings of the 29th ACM SIGKDD Conference on Knowledge Discovery and Data Mining}}. \bibinfo{pages}{262--273}.
\newblock


\bibitem[Chien et~al\mbox{.}(2020)]%
        {chien2020adaptive}
\bibfield{author}{\bibinfo{person}{Eli Chien}, \bibinfo{person}{Jianhao Peng}, \bibinfo{person}{Pan Li}, {and} \bibinfo{person}{Olgica Milenkovic}.} \bibinfo{year}{2020}\natexlab{}.
\newblock \showarticletitle{Adaptive universal generalized pagerank graph neural network}.
\newblock \bibinfo{journal}{\emph{arXiv preprint arXiv:2006.07988}} (\bibinfo{year}{2020}).
\newblock


\bibitem[Daubechies et~al\mbox{.}(2004)]%
        {daubechies2004iterative}
\bibfield{author}{\bibinfo{person}{Ingrid Daubechies}, \bibinfo{person}{Michel Defrise}, {and} \bibinfo{person}{Christine De~Mol}.} \bibinfo{year}{2004}\natexlab{}.
\newblock \showarticletitle{An iterative thresholding algorithm for linear inverse problems with a sparsity constraint}.
\newblock \bibinfo{journal}{\emph{Communications on Pure and Applied Mathematics: A Journal Issued by the Courant Institute of Mathematical Sciences}} \bibinfo{volume}{57}, \bibinfo{number}{11} (\bibinfo{year}{2004}), \bibinfo{pages}{1413--1457}.
\newblock


\bibitem[Fountoulakis et~al\mbox{.}(2019)]%
        {fountoulakis2019variational}
\bibfield{author}{\bibinfo{person}{Kimon Fountoulakis}, \bibinfo{person}{Farbod Roosta-Khorasani}, \bibinfo{person}{Julian Shun}, \bibinfo{person}{Xiang Cheng}, {and} \bibinfo{person}{Michael~W Mahoney}.} \bibinfo{year}{2019}\natexlab{}.
\newblock \showarticletitle{Variational perspective on local graph clustering}.
\newblock \bibinfo{journal}{\emph{Mathematical Programming}}  \bibinfo{volume}{174} (\bibinfo{year}{2019}), \bibinfo{pages}{553--573}.
\newblock


\bibitem[Fountoulakis and Yang(2022)]%
        {fountoulakis2022open}
\bibfield{author}{\bibinfo{person}{Kimon Fountoulakis} {and} \bibinfo{person}{Shenghao Yang}.} \bibinfo{year}{2022}\natexlab{}.
\newblock \showarticletitle{Open Problem: Running time complexity of accelerated l-1 regularized PageRank}. In \bibinfo{booktitle}{\emph{Conference on Learning Theory}}. PMLR, \bibinfo{pages}{5630--5632}.
\newblock


\bibitem[Fu and He(2021)]%
        {fu2021sdg}
\bibfield{author}{\bibinfo{person}{Dongqi Fu} {and} \bibinfo{person}{Jingrui He}.} \bibinfo{year}{2021}\natexlab{}.
\newblock \showarticletitle{SDG: a simplified and dynamic graph neural network}. In \bibinfo{booktitle}{\emph{Proceedings of the 44th International ACM SIGIR Conference on Research and Development in Information Retrieval}}. \bibinfo{pages}{2273--2277}.
\newblock


\bibitem[Gasteiger et~al\mbox{.}(2018)]%
        {gasteiger2018predict}
\bibfield{author}{\bibinfo{person}{Johannes Gasteiger}, \bibinfo{person}{Aleksandar Bojchevski}, {and} \bibinfo{person}{Stephan G{\"u}nnemann}.} \bibinfo{year}{2018}\natexlab{}.
\newblock \showarticletitle{Predict then propagate: Graph neural networks meet personalized pagerank}.
\newblock \bibinfo{journal}{\emph{arXiv preprint arXiv:1810.05997}} (\bibinfo{year}{2018}).
\newblock


\bibitem[Guo et~al\mbox{.}(2019)]%
        {guo2019inferring}
\bibfield{author}{\bibinfo{person}{Xingzhi Guo}, \bibinfo{person}{Yu-Cian Huang}, \bibinfo{person}{Edwinn Gamborino}, \bibinfo{person}{Shih-Huan Tseng}, \bibinfo{person}{Li-Chen Fu}, {and} \bibinfo{person}{Su-Ling Yeh}.} \bibinfo{year}{2019}\natexlab{}.
\newblock \showarticletitle{Inferring human feelings and desires for human-robot trust promotion}. In \bibinfo{booktitle}{\emph{International Conference on Human-Computer Interaction}}. Springer, \bibinfo{pages}{365--375}.
\newblock


\bibitem[Guo et~al\mbox{.}(2022a)]%
        {guo2022verba}
\bibfield{author}{\bibinfo{person}{Xingzhi Guo}, \bibinfo{person}{Brian Kondracki}, \bibinfo{person}{Nick Nikiforakis}, {and} \bibinfo{person}{Steven Skiena}.} \bibinfo{year}{2022}\natexlab{a}.
\newblock \showarticletitle{Verba volant, scripta volant: Understanding post-publication title changes in news outlets}. In \bibinfo{booktitle}{\emph{Proceedings of the ACM Web Conference 2022}}. \bibinfo{pages}{588--598}.
\newblock


\bibitem[Guo and Skiena(2022)]%
        {guo2022hierarchies}
\bibfield{author}{\bibinfo{person}{Xingzhi Guo} {and} \bibinfo{person}{Steven Skiena}.} \bibinfo{year}{2022}\natexlab{}.
\newblock \showarticletitle{Hierarchies over Vector Space: Orienting Word and Graph Embeddings}.
\newblock \bibinfo{journal}{\emph{arXiv preprint arXiv:2211.01430}} (\bibinfo{year}{2022}).
\newblock


\bibitem[Guo et~al\mbox{.}(2021)]%
        {guo2021subset}
\bibfield{author}{\bibinfo{person}{Xingzhi Guo}, \bibinfo{person}{Baojian Zhou}, {and} \bibinfo{person}{Steven Skiena}.} \bibinfo{year}{2021}\natexlab{}.
\newblock \showarticletitle{Subset Node Representation Learning over Large Dynamic Graphs}. In \bibinfo{booktitle}{\emph{Proceedings of the 27th ACM SIGKDD Conference on Knowledge Discovery \& Data Mining}}. \bibinfo{pages}{516--526}.
\newblock


\bibitem[Guo et~al\mbox{.}(2022b)]%
        {guo2022subset}
\bibfield{author}{\bibinfo{person}{Xingzhi Guo}, \bibinfo{person}{Baojian Zhou}, {and} \bibinfo{person}{Steven Skiena}.} \bibinfo{year}{2022}\natexlab{b}.
\newblock \showarticletitle{Subset Node Anomaly Tracking over Large Dynamic Graphs}. In \bibinfo{booktitle}{\emph{Proceedings of the 28th ACM SIGKDD Conference on Knowledge Discovery and Data Mining}}. \bibinfo{pages}{475--485}.
\newblock


\bibitem[Hajiramezanali et~al\mbox{.}(2019)]%
        {hajiramezanali2019variational}
\bibfield{author}{\bibinfo{person}{Ehsan Hajiramezanali}, \bibinfo{person}{Arman Hasanzadeh}, \bibinfo{person}{Krishna Narayanan}, \bibinfo{person}{Nick Duffield}, \bibinfo{person}{Mingyuan Zhou}, {and} \bibinfo{person}{Xiaoning Qian}.} \bibinfo{year}{2019}\natexlab{}.
\newblock \showarticletitle{Variational graph recurrent neural networks}.
\newblock \bibinfo{journal}{\emph{Advances in neural information processing systems}}  \bibinfo{volume}{32} (\bibinfo{year}{2019}).
\newblock


\bibitem[Hamilton et~al\mbox{.}(2017)]%
        {hamilton2017inductive}
\bibfield{author}{\bibinfo{person}{Will Hamilton}, \bibinfo{person}{Zhitao Ying}, {and} \bibinfo{person}{Jure Leskovec}.} \bibinfo{year}{2017}\natexlab{}.
\newblock \showarticletitle{Inductive representation learning on large graphs}.
\newblock \bibinfo{journal}{\emph{Advances in neural information processing systems}}  \bibinfo{volume}{30} (\bibinfo{year}{2017}).
\newblock


\bibitem[Kim and Oh(2022)]%
        {kim2022find}
\bibfield{author}{\bibinfo{person}{Dongkwan Kim} {and} \bibinfo{person}{Alice Oh}.} \bibinfo{year}{2022}\natexlab{}.
\newblock \showarticletitle{How to find your friendly neighborhood: Graph attention design with self-supervision}.
\newblock \bibinfo{journal}{\emph{arXiv preprint arXiv:2204.04879}} (\bibinfo{year}{2022}).
\newblock


\bibitem[Kipf and Welling(2016)]%
        {kipf2016semi}
\bibfield{author}{\bibinfo{person}{Thomas~N Kipf} {and} \bibinfo{person}{Max Welling}.} \bibinfo{year}{2016}\natexlab{}.
\newblock \showarticletitle{Semi-supervised classification with graph convolutional networks}.
\newblock \bibinfo{journal}{\emph{arXiv preprint arXiv:1609.02907}} (\bibinfo{year}{2016}).
\newblock


\bibitem[Kumar et~al\mbox{.}(2019)]%
        {kumar2019predicting}
\bibfield{author}{\bibinfo{person}{Srijan Kumar}, \bibinfo{person}{Xikun Zhang}, {and} \bibinfo{person}{Jure Leskovec}.} \bibinfo{year}{2019}\natexlab{}.
\newblock \showarticletitle{Predicting dynamic embedding trajectory in temporal interaction networks}. In \bibinfo{booktitle}{\emph{Proceedings of the 25th ACM SIGKDD Conference on Knowledge Discovery \& Data Mining}}. \bibinfo{pages}{1269--1278}.
\newblock


\bibitem[Lin et~al\mbox{.}(2023)]%
        {lin2023comet}
\bibfield{author}{\bibinfo{person}{Zhuoyi Lin}, \bibinfo{person}{Lei Feng}, \bibinfo{person}{Xingzhi Guo}, \bibinfo{person}{Yu Zhang}, \bibinfo{person}{Rui Yin}, \bibinfo{person}{Chee~Keong Kwoh}, {and} \bibinfo{person}{Chi Xu}.} \bibinfo{year}{2023}\natexlab{}.
\newblock \showarticletitle{Comet: Convolutional dimension interaction for collaborative filtering}.
\newblock \bibinfo{journal}{\emph{ACM Transactions on Intelligent Systems and Technology}} \bibinfo{volume}{14}, \bibinfo{number}{4} (\bibinfo{year}{2023}), \bibinfo{pages}{1--18}.
\newblock


\bibitem[Liu et~al\mbox{.}(2021)]%
        {liu2021graph}
\bibfield{author}{\bibinfo{person}{Xiaorui Liu}, \bibinfo{person}{Jiayuan Ding}, \bibinfo{person}{Wei Jin}, \bibinfo{person}{Han Xu}, \bibinfo{person}{Yao Ma}, \bibinfo{person}{Zitao Liu}, {and} \bibinfo{person}{Jiliang Tang}.} \bibinfo{year}{2021}\natexlab{}.
\newblock \showarticletitle{Graph neural networks with adaptive residual}.
\newblock \bibinfo{journal}{\emph{Advances in Neural Information Processing Systems}}  \bibinfo{volume}{34} (\bibinfo{year}{2021}), \bibinfo{pages}{9720--9733}.
\newblock


\bibitem[McPherson et~al\mbox{.}(2001)]%
        {mcpherson2001birds}
\bibfield{author}{\bibinfo{person}{Miller McPherson}, \bibinfo{person}{Lynn Smith-Lovin}, {and} \bibinfo{person}{James~M Cook}.} \bibinfo{year}{2001}\natexlab{}.
\newblock \showarticletitle{Birds of a feather: Homophily in social networks}.
\newblock \bibinfo{journal}{\emph{Annual review of sociology}} \bibinfo{volume}{27}, \bibinfo{number}{1} (\bibinfo{year}{2001}), \bibinfo{pages}{415--444}.
\newblock


\bibitem[Nutini et~al\mbox{.}(2017)]%
        {nutini2017let}
\bibfield{author}{\bibinfo{person}{Julie Nutini}, \bibinfo{person}{Issam Laradji}, {and} \bibinfo{person}{Mark Schmidt}.} \bibinfo{year}{2017}\natexlab{}.
\newblock \showarticletitle{Let's Make Block Coordinate Descent Converge Faster: Faster Greedy Rules, Message-Passing, Active-Set Complexity, and Superlinear Convergence}.
\newblock \bibinfo{journal}{\emph{arXiv preprint arXiv:1712.08859}} (\bibinfo{year}{2017}).
\newblock


\bibitem[Oono and Suzuki(2020)]%
        {Oono2020Graph}
\bibfield{author}{\bibinfo{person}{Kenta Oono} {and} \bibinfo{person}{Taiji Suzuki}.} \bibinfo{year}{2020}\natexlab{}.
\newblock \showarticletitle{Graph Neural Networks Exponentially Lose Expressive Power for Node Classification}. In \bibinfo{booktitle}{\emph{International Conference on Learning Representations}}.
\newblock
\urldef\tempurl%
\url{https://openreview.net/forum?id=S1ldO2EFPr}
\showURL{%
\tempurl}


\bibitem[Page et~al\mbox{.}(1999)]%
        {page1999pagerank}
\bibfield{author}{\bibinfo{person}{Lawrence Page}, \bibinfo{person}{Sergey Brin}, \bibinfo{person}{Rajeev Motwani}, {and} \bibinfo{person}{Terry Winograd}.} \bibinfo{year}{1999}\natexlab{}.
\newblock \bibinfo{booktitle}{\emph{The PageRank citation ranking: Bringing order to the web.}}
\newblock \bibinfo{type}{{T}echnical {R}eport}. \bibinfo{institution}{Stanford InfoLab}.
\newblock


\bibitem[Post{\u{a}}varu et~al\mbox{.}(2020)]%
        {postuavaru2020instantembedding}
\bibfield{author}{\bibinfo{person}{{\c{S}}tefan Post{\u{a}}varu}, \bibinfo{person}{Anton Tsitsulin}, \bibinfo{person}{Filipe Miguel~Gon{\c{c}}alves de Almeida}, \bibinfo{person}{Yingtao Tian}, \bibinfo{person}{Silvio Lattanzi}, {and} \bibinfo{person}{Bryan Perozzi}.} \bibinfo{year}{2020}\natexlab{}.
\newblock \showarticletitle{InstantEmbedding: Efficient Local Node Representations}.
\newblock \bibinfo{journal}{\emph{arXiv preprint arXiv:2010.06992}} (\bibinfo{year}{2020}).
\newblock


\bibitem[Rossi et~al\mbox{.}(2020)]%
        {rossi2020temporal}
\bibfield{author}{\bibinfo{person}{Emanuele Rossi}, \bibinfo{person}{Ben Chamberlain}, \bibinfo{person}{Fabrizio Frasca}, \bibinfo{person}{Davide Eynard}, \bibinfo{person}{Federico Monti}, {and} \bibinfo{person}{Michael Bronstein}.} \bibinfo{year}{2020}\natexlab{}.
\newblock \showarticletitle{Temporal graph networks for deep learning on dynamic graphs}.
\newblock \bibinfo{journal}{\emph{arXiv preprint arXiv:2006.10637}} (\bibinfo{year}{2020}).
\newblock


\bibitem[Sankar et~al\mbox{.}(2020)]%
        {sankar2020dysat}
\bibfield{author}{\bibinfo{person}{Aravind Sankar}, \bibinfo{person}{Yanhong Wu}, \bibinfo{person}{Liang Gou}, \bibinfo{person}{Wei Zhang}, {and} \bibinfo{person}{Hao Yang}.} \bibinfo{year}{2020}\natexlab{}.
\newblock \showarticletitle{Dysat: Deep neural representation learning on dynamic graphs via self-attention networks}. In \bibinfo{booktitle}{\emph{ACM International Conference on Web Search and Data Mining}}. \bibinfo{pages}{519--527}.
\newblock


\bibitem[Sultan et~al\mbox{.}(2022)]%
        {sultan2022low}
\bibfield{author}{\bibinfo{person}{Syed~Fahad Sultan}, \bibinfo{person}{Xingzhi Guo}, {and} \bibinfo{person}{Steven Skiena}.} \bibinfo{year}{2022}\natexlab{}.
\newblock \showarticletitle{Low-dimensional genotype embeddings for predictive models}. In \bibinfo{booktitle}{\emph{Proceedings of the 13th ACM International Conference on Bioinformatics, Computational Biology and Health Informatics}}. \bibinfo{pages}{1--4}.
\newblock


\bibitem[Tibshirani(1996)]%
        {tibshirani1996regression}
\bibfield{author}{\bibinfo{person}{Robert Tibshirani}.} \bibinfo{year}{1996}\natexlab{}.
\newblock \showarticletitle{Regression shrinkage and selection via the lasso}.
\newblock \bibinfo{journal}{\emph{Journal of the Royal Statistical Society: Series B (Methodological)}} \bibinfo{volume}{58}, \bibinfo{number}{1} (\bibinfo{year}{1996}), \bibinfo{pages}{267--288}.
\newblock


\bibitem[Trivedi et~al\mbox{.}(2019)]%
        {trivedi2019dyrep}
\bibfield{author}{\bibinfo{person}{Rakshit Trivedi}, \bibinfo{person}{Mehrdad Farajtabar}, \bibinfo{person}{Prasenjeet Biswal}, {and} \bibinfo{person}{Hongyuan Zha}.} \bibinfo{year}{2019}\natexlab{}.
\newblock \showarticletitle{Dyrep: Learning representations over dynamic graphs}. In \bibinfo{booktitle}{\emph{The International Conference on Learning Representations (ICLR)}}.
\newblock


\bibitem[Vaswani et~al\mbox{.}(2017)]%
        {vaswani2017attention}
\bibfield{author}{\bibinfo{person}{Ashish Vaswani}, \bibinfo{person}{Noam Shazeer}, \bibinfo{person}{Niki Parmar}, \bibinfo{person}{Jakob Uszkoreit}, \bibinfo{person}{Llion Jones}, \bibinfo{person}{Aidan~N Gomez}, \bibinfo{person}{{\L}ukasz Kaiser}, {and} \bibinfo{person}{Illia Polosukhin}.} \bibinfo{year}{2017}\natexlab{}.
\newblock \showarticletitle{Attention is all you need}.
\newblock \bibinfo{journal}{\emph{Advances in neural information processing systems}}  \bibinfo{volume}{30} (\bibinfo{year}{2017}).
\newblock


\bibitem[Wang et~al\mbox{.}(2021)]%
        {wang2021approximate}
\bibfield{author}{\bibinfo{person}{Hanzhi Wang}, \bibinfo{person}{Mingguo He}, \bibinfo{person}{Zhewei Wei}, \bibinfo{person}{Sibo Wang}, \bibinfo{person}{Ye Yuan}, \bibinfo{person}{Xiaoyong Du}, {and} \bibinfo{person}{Ji-Rong Wen}.} \bibinfo{year}{2021}\natexlab{}.
\newblock \showarticletitle{Approximate graph propagation}. In \bibinfo{booktitle}{\emph{Proceedings of the 27th ACM SIGKDD Conference on Knowledge Discovery \& Data Mining}}. \bibinfo{pages}{1686--1696}.
\newblock


\bibitem[Wang et~al\mbox{.}(2022b)]%
        {wang2022edge}
\bibfield{author}{\bibinfo{person}{Hanzhi Wang}, \bibinfo{person}{Zhewei Wei}, \bibinfo{person}{Junhao Gan}, \bibinfo{person}{Ye Yuan}, \bibinfo{person}{Xiaoyong Du}, {and} \bibinfo{person}{Ji-Rong Wen}.} \bibinfo{year}{2022}\natexlab{b}.
\newblock \showarticletitle{Edge-based local push for personalized PageRank}.
\newblock \bibinfo{journal}{\emph{arXiv preprint arXiv:2203.07937}} (\bibinfo{year}{2022}).
\newblock


\bibitem[Wang(2023)]%
        {wang2023knowledge2}
\bibfield{author}{\bibinfo{person}{Yuheng Wang}.} \bibinfo{year}{2023}\natexlab{}.
\newblock \emph{\bibinfo{title}{Knowledge Authoring With Factual English, Rules, and Actions}}.
\newblock \bibinfo{thesistype}{Ph.\,D. Dissertation}. \bibinfo{school}{State University of New York at Stony Brook}.
\newblock


\bibitem[Wang et~al\mbox{.}(2022a)]%
        {wang2022knowledge}
\bibfield{author}{\bibinfo{person}{Yuheng Wang}, \bibinfo{person}{Giorgian Borca-Tasciuc}, \bibinfo{person}{Nikhil Goel}, \bibinfo{person}{Paul Fodor}, {and} \bibinfo{person}{Michael Kifer}.} \bibinfo{year}{2022}\natexlab{a}.
\newblock \showarticletitle{Knowledge authoring with factual english}.
\newblock \bibinfo{journal}{\emph{arXiv preprint arXiv:2208.03094}} (\bibinfo{year}{2022}).
\newblock


\bibitem[Wang et~al\mbox{.}(2023)]%
        {wang2023knowledge}
\bibfield{author}{\bibinfo{person}{Yuheng Wang}, \bibinfo{person}{Paul Fodor}, {and} \bibinfo{person}{Michael Kifer}.} \bibinfo{year}{2023}\natexlab{}.
\newblock \showarticletitle{Knowledge Authoring for Rules and Actions}.
\newblock \bibinfo{journal}{\emph{Theory and Practice of Logic Programming}} \bibinfo{volume}{23}, \bibinfo{number}{4} (\bibinfo{year}{2023}), \bibinfo{pages}{797--811}.
\newblock


\bibitem[Wen et~al\mbox{.}(2022)]%
        {wen2022disentangled}
\bibfield{author}{\bibinfo{person}{Qianlong Wen}, \bibinfo{person}{Zhongyu Ouyang}, \bibinfo{person}{Jianfei Zhang}, \bibinfo{person}{Yiyue Qian}, \bibinfo{person}{Yanfang Ye}, {and} \bibinfo{person}{Chuxu Zhang}.} \bibinfo{year}{2022}\natexlab{}.
\newblock \showarticletitle{Disentangled dynamic heterogeneous graph learning for opioid overdose prediction}. In \bibinfo{booktitle}{\emph{Proceedings of the 28th ACM SIGKDD Conference on Knowledge Discovery and Data Mining}}. \bibinfo{pages}{2009--2019}.
\newblock


\bibitem[Wu et~al\mbox{.}(2019)]%
        {wu2019simplifying}
\bibfield{author}{\bibinfo{person}{Felix Wu}, \bibinfo{person}{Amauri Souza}, \bibinfo{person}{Tianyi Zhang}, \bibinfo{person}{Christopher Fifty}, \bibinfo{person}{Tao Yu}, {and} \bibinfo{person}{Kilian Weinberger}.} \bibinfo{year}{2019}\natexlab{}.
\newblock \showarticletitle{Simplifying graph convolutional networks}. In \bibinfo{booktitle}{\emph{International conference on machine learning}}. PMLR, \bibinfo{pages}{6861--6871}.
\newblock


\bibitem[Wu et~al\mbox{.}(2021)]%
        {wu2021unifying}
\bibfield{author}{\bibinfo{person}{Hao Wu}, \bibinfo{person}{Junhao Gan}, \bibinfo{person}{Zhewei Wei}, {and} \bibinfo{person}{Rui Zhang}.} \bibinfo{year}{2021}\natexlab{}.
\newblock \showarticletitle{Unifying the Global and Local Approaches: An Efficient Power Iteration with Forward Push}. In \bibinfo{booktitle}{\emph{Proceedings of the 2021 International Conference on Management of Data}}. \bibinfo{pages}{1996--2008}.
\newblock


\bibitem[Xu et~al\mbox{.}(2020)]%
        {xu2020inductive}
\bibfield{author}{\bibinfo{person}{Da Xu}, \bibinfo{person}{Chuanwei Ruan}, \bibinfo{person}{Evren Korpeoglu}, \bibinfo{person}{Sushant Kumar}, {and} \bibinfo{person}{Kannan Achan}.} \bibinfo{year}{2020}\natexlab{}.
\newblock \showarticletitle{Inductive representation learning on temporal graphs}.
\newblock \bibinfo{journal}{\emph{arXiv preprint arXiv:2002.07962}} (\bibinfo{year}{2020}).
\newblock


\bibitem[Yu et~al\mbox{.}(2018)]%
        {yu2018netwalk}
\bibfield{author}{\bibinfo{person}{Wenchao Yu}, \bibinfo{person}{Wei Cheng}, \bibinfo{person}{Charu~C Aggarwal}, \bibinfo{person}{Kai Zhang}, \bibinfo{person}{Haifeng Chen}, {and} \bibinfo{person}{Wei Wang}.} \bibinfo{year}{2018}\natexlab{}.
\newblock \showarticletitle{Netwalk: A flexible deep embedding approach for anomaly detection in dynamic networks}. In \bibinfo{booktitle}{\emph{Proceedings of the 24th ACM SIGKDD Conference on Knowledge Discovery \& Data Mining}}. \bibinfo{pages}{2672--2681}.
\newblock


\bibitem[Zhang et~al\mbox{.}(2023)]%
        {zhang2023subanom}
\bibfield{author}{\bibinfo{person}{Chi Zhang}, \bibinfo{person}{Wenkai Xiang}, \bibinfo{person}{Xingzhi Guo}, \bibinfo{person}{Baojian Zhou}, {and} \bibinfo{person}{Deqing Yang}.} \bibinfo{year}{2023}\natexlab{}.
\newblock \showarticletitle{SubAnom: Efficient Subgraph Anomaly Detection Framework over Dynamic Graphs}. In \bibinfo{booktitle}{\emph{2023 IEEE International Conference on Data Mining Workshops (ICDMW)}}. IEEE, \bibinfo{pages}{1178--1185}.
\newblock


\bibitem[Zhang et~al\mbox{.}(2016a)]%
        {zhang2016approximate}
\bibfield{author}{\bibinfo{person}{Hongyang Zhang}, \bibinfo{person}{Peter Lofgren}, {and} \bibinfo{person}{Ashish Goel}.} \bibinfo{year}{2016}\natexlab{a}.
\newblock \showarticletitle{Approximate personalized pagerank on dynamic graphs}. In \bibinfo{booktitle}{\emph{Proceedings of the 22nd ACM SIGKDD Conference on Knowledge Discovery \& Data Mining}}. \bibinfo{pages}{1315--1324}.
\newblock


\bibitem[Zhang et~al\mbox{.}(2016b)]%
        {zhang2016symmetrical}
\bibfield{author}{\bibinfo{person}{Junqi Zhang}, \bibinfo{person}{Yuheng Wang}, \bibinfo{person}{Cheng Wang}, {and} \bibinfo{person}{Mengchu Zhou}.} \bibinfo{year}{2016}\natexlab{b}.
\newblock \showarticletitle{Symmetrical hierarchical stochastic searching on the line in informative and deceptive environments}.
\newblock \bibinfo{journal}{\emph{IEEE transactions on cybernetics}} \bibinfo{volume}{47}, \bibinfo{number}{3} (\bibinfo{year}{2016}), \bibinfo{pages}{626--635}.
\newblock


\bibitem[Zhang et~al\mbox{.}(2017b)]%
        {zhang2017fast}
\bibfield{author}{\bibinfo{person}{Junqi Zhang}, \bibinfo{person}{Yuheng Wang}, \bibinfo{person}{Cheng Wang}, {and} \bibinfo{person}{MengChu Zhou}.} \bibinfo{year}{2017}\natexlab{b}.
\newblock \showarticletitle{Fast variable structure stochastic automaton for discovering and tracking spatiotemporal event patterns}.
\newblock \bibinfo{journal}{\emph{IEEE transactions on cybernetics}} \bibinfo{volume}{48}, \bibinfo{number}{3} (\bibinfo{year}{2017}), \bibinfo{pages}{890--903}.
\newblock


\bibitem[Zhang et~al\mbox{.}(2017a)]%
        {zhang2017fast2}
\bibfield{author}{\bibinfo{person}{JunQi Zhang}, \bibinfo{person}{YuHeng Wang}, {and} \bibinfo{person}{MengChu Zhou}.} \bibinfo{year}{2017}\natexlab{a}.
\newblock \showarticletitle{Fast adaptive search on the line in dual environments}. In \bibinfo{booktitle}{\emph{2017 13th IEEE Conference on Automation Science and Engineering (CASE)}}. IEEE, \bibinfo{pages}{1540--1545}.
\newblock


\bibitem[Zhang et~al\mbox{.}(2018)]%
        {zhang2018dual}
\bibfield{author}{\bibinfo{person}{Junqi Zhang}, \bibinfo{person}{Xixun Zhu}, \bibinfo{person}{Yuheng Wang}, {and} \bibinfo{person}{MengChu Zhou}.} \bibinfo{year}{2018}\natexlab{}.
\newblock \showarticletitle{Dual-environmental particle swarm optimizer in noisy and noise-free environments}.
\newblock \bibinfo{journal}{\emph{IEEE transactions on cybernetics}} \bibinfo{volume}{49}, \bibinfo{number}{6} (\bibinfo{year}{2018}), \bibinfo{pages}{2011--2021}.
\newblock


\bibitem[Zheng et~al\mbox{.}(2022)]%
        {zheng2022instant}
\bibfield{author}{\bibinfo{person}{Yanping Zheng}, \bibinfo{person}{Hanzhi Wang}, \bibinfo{person}{Zhewei Wei}, \bibinfo{person}{Jiajun Liu}, {and} \bibinfo{person}{Sibo Wang}.} \bibinfo{year}{2022}\natexlab{}.
\newblock \showarticletitle{Instant Graph Neural Networks for Dynamic Graphs}. In \bibinfo{booktitle}{\emph{Proceedings of the 28th ACM SIGKDD Conference on Knowledge Discovery and Data Mining}}. \bibinfo{pages}{2605--2615}.
\newblock


\bibitem[Zhou et~al\mbox{.}(2019)]%
        {zhou2019dual}
\bibfield{author}{\bibinfo{person}{Baojian Zhou}, \bibinfo{person}{Feng Chen}, {and} \bibinfo{person}{Yiming Ying}.} \bibinfo{year}{2019}\natexlab{}.
\newblock \showarticletitle{Dual averaging method for online graph-structured sparsity}. In \bibinfo{booktitle}{\emph{Proceedings of the 25th ACM SIGKDD International Conference on Knowledge Discovery \& Data Mining}}. \bibinfo{pages}{436--446}.
\newblock


\bibitem[Zhou et~al\mbox{.}(2024)]%
        {zhou2024iterative}
\bibfield{author}{\bibinfo{person}{Baojian Zhou}, \bibinfo{person}{Yifan Sun}, \bibinfo{person}{Reza~Babanezhad Harikandeh}, \bibinfo{person}{Xingzhi Guo}, \bibinfo{person}{Deqing Yang}, {and} \bibinfo{person}{Yanghua Xiao}.} \bibinfo{year}{2024}\natexlab{}.
\newblock \showarticletitle{Iterative Methods via Locally Evolving Set Process}.
\newblock \bibinfo{journal}{\emph{arXiv preprint arXiv:2410.15020}} (\bibinfo{year}{2024}).
\newblock


\end{thebibliography}
\clearpage

\end{document}